\documentclass[runningheads]{llncs}

\usepackage[T1]{fontenc}
\usepackage{times}
\usepackage{soul}
\usepackage[utf8]{inputenc}
\usepackage{graphicx}
\usepackage{amsmath}
\usepackage{amssymb}

\usepackage{amsthm}
\usepackage{booktabs}
\usepackage{algorithm} 
\usepackage[switch]{lineno}

\usepackage{stmaryrd}

\usepackage{algpseudocode}
\usepackage{multirow}
\usepackage{multicol}
\usepackage{subcaption}
\usepackage{url}
\usepackage{xspace}
\usepackage{lscape}
\newcommand{\approach}{\textsc{Ebr}\xspace}

\author{
Louis Mozart Kamdem Teyou
\and
Caglar Demir \And
Axel-Cyrille Ngonga Ngomo\\
\affiliations
$^1$Heinz Nixdorf Institute\\
$^2$Paderborn University\\
\emails
\{louis888, cdmir, axel.ngonga\}@upb.de
}

\begin{document}
\title{Semantics-Aware Caching for Concept Learning}

\author{Louis Mozart Kamdem Teyou
\inst{1,2}\orcidID{0000-0001-7975-8794} \and
Caglar Demir\inst{1,2}\orcidID{0000-0001-8970-3850} \and
Axel-Cyrille Ngonga Ngomo\inst{1,2}\orcidID{0000-0001-7112-3516}
\email
\{louis888, cdmir, axel.ngonga\}@upb.de}

\authorrunning{Louis Mozart et al.}

\institute{Heinz Nixdorf Institute, Furstenalle 11,  33102 Paderborn, Germany \and
Paderborn University, Warburgerstr. 100, 33098 Paderborn, Germany}

\maketitle              

\begin{abstract}
 Concept learning is a form of supervised machine learning that operates on knowledge bases in description logics.  
State-of-the-art concept learners 
often rely on an iterative search through a countably infinite concept space. In each iteration, they retrieve instances of candidate solutions to select the best concept for the next iteration. 
While simple learning problems might require a few dozen concept retrieval calls to find a suitable solution, complex learning problems might require thousands of calls, making concept retrieval a major runtime bottleneck. We alleviate the resulting runtime challenge by presenting a semantics-aware caching approach. Our cache is essentially a subsumption-aware map that links concepts to a set of instances via crisp set operations. 
We evaluate our approach on four datasets using four symbolic reasoners, one neuro-symbolic reasoner, and five cache eviction policies. The results show that our cache reduces concept retrieval and concept learning runtimes by up to an order of magnitude while remaining effective across both symbolic and neuro-symbolic reasoning reasoners. 
\end{abstract}

\section{Introduction}

Knowledge bases have become first-class citizens of the Web, with roughly 50\% of websites containing structured data expressed in RDF knowledge bases.\footnote{\url{https://webdatacommons.org/structureddata/}} 
Given the advent of machine learning across disciplines and on corpora of increasing size, it stands to reason that the over $10^{11}$ assertions available on the Web need to be made amenable to machine learning. Embedding approaches and geometric deep learning are used to map the constituents of knowledge bases to vectorial representations \cite{DBLP:journals/symmetry/WangQW21}. 

While effective for many predictive tasks, these methods typically approximate the relational structure of knowledge bases and do not explicitly preserve the logical semantics of RDF and description logics. In contrast, concept learning (also called class expression learning, short: CEL) \cite{fanizzi2018dlfoil,rizzo2020class,DBLP:conf/pkdd/KaralisBDHN24} operates directly on symbolic representations and produces interpretable models in the form of description logic class expressions.

The goal of CEL is to learn a class expression that describes a set of positive examples while excluding negative ones \cite{lehmann2010concept,lehmann2011class}. In practice, this requires searching in an infinite, quasi-ordered hypothesis space defined by the expressivity of the underlying description logic (short: DL). Most state-of-the-art CEL algorithms use a top-down refinement operator that starts from the most general concept (e.g., $\top$) and iteratively specializes it through a sequence of refinement steps \cite{lehmann2010concept}. This iterative process continues until a hypothesis that best approximates the target concept in terms of covering positive examples while avoiding negatives is reached. 

CEL has demonstrated its effectiveness across various application domains. In software engineering, it has been employed to support the development of software information systems by facilitating semantic integration and information retrieval \cite{nardi2003introduction}. In ontology engineering, CEL contributes to the (semi-)automatic construction, refinement, and validation of ontologies by learning new concept definitions from instance data \cite{lehmann2010concept}. In the medical field, CEL has been applied to tasks such as the prediction of protein functions, where logical definitions of protein classes enhance interpretability and biological insight \cite{kulmanov2018deepgo}. Additionally, it has been used in content recommendation systems, where learned class expressions help model user preferences and content categories in a transparent and semantically meaningful manner \cite{oramas2016sound}.

A central computational bottleneck of CEL is the repeated use of description logic reasoners to evaluate candidate class expressions. Each candidate must be submitted to a reasoner to retrieve its instances, an operation that can be computationally expensive. During the search process, hundreds or even thousands of candidate expressions may be evaluated, resulting in many repeated instance-retrieval calls. Consequently, the efficiency of concept retrieval has a direct impact on the overall runtime of concept learning algorithms.

Despite the importance of this operation, we could not identify prior work that systematically investigates caching mechanisms for accelerating concept retrieval. In this work, we address this bottleneck by introducing a semantics-aware caching mechanism designed to accelerate concept retrieval in description logic reasoners. Our cache can be viewed as a subsumption-aware map that associates class expressions with their corresponding sets of instances. By exploiting the semantics of $\mathcal{ALC}$, the cache can derive the instances of new concepts from previously computed ones through simple set operations, thereby avoiding redundant reasoning calls.

Traditional approaches such as ELK \cite{kazakov2014incredible} demonstrate how materialization-based techniques can significantly improve reasoning efficiency by precomputing subsumptions and class memberships. Although our work shares the general goal of improving scalability, the two approaches differ fundamentally. ELK is an optimized reasoner designed specifically for the $\mathcal{EL}^{++}$ description logic. In contrast, our approach does not introduce a new reasoner. Instead, it provides a complementary caching layer that accelerates concept retrieval for existing reasoners without modifying their internal algorithms. For this reason, a direct experimental comparison with ELK is not meaningful.

We evaluate the proposed caching mechanism on four datasets using four symbolic reasoners and one neuro-symbolic reasoner. Experiments are conducted for both concept retrieval and full concept-learning pipelines. Our results show that the proposed cache can substantially improve performance, reducing concept retrieval time by up to $80\%$ for slower reasoners and by up to $20\%$ for faster ones when sufficient cache capacity is available. When integrated into concept learning systems, the cache can reduce overall learning runtimes by up to three orders of magnitude.

Our code, datasets, and experimental setup are available in the supplementary material.\footnote{\url{https://github.com/Louis-Mozart/Cache-algorithm-for-concept-learners}}

 \section{Background}
 \label{sec:related_work}

 \subsection{Concept Learning Algorithms}

CEL algorithms have significantly evolved during the last decade. Among the early approaches, CELOE \cite{lehmann2011class}, OCEL, and ELTL \cite{buhmann2016dl} are foundational examples of search-based CEL methods. These algorithms rely on refinement operators to iteratively generate and evaluate candidate class expressions and are all implemented into the DL-Learner framework \cite{lehmann2010concept}. OCEL uses heuristic rules to guide the search process, reducing the exploration of irrelevant or overly complex expressions. CELOE extends upon OCEL by improving the heuristic function, prioritizing syntactically shorter and semantically relevant expressions, which has made it one of the most effective search-based CEL algorithms within the DL-Learner framework \cite{kouagou2023neural}.

EvoLearner \cite{heindorf2022evolearner}, a more recent development, takes an evolutionary approach to CEL. Instead of relying only on refinement operators, EvoLearner initializes its population of candidate expressions through random walks on the knowledge graph, which are then converted into description logic concepts represented as abstract syntax trees. These candidate concepts are refined further using mutation and crossover operations, enabling EvoLearner to outperform traditional refinement operator-based methods like CELOE and OCEL in both efficiency and accuracy.  Another example is ECII \cite{sarker2019efficient}, a search-based algorithm that does not employ refinement operators, invoking a reasoner only once during execution. This design choice allows ECII to bypass the iterative computational overhead that is common in traditional methods.


As ontologies grow increasingly complex and data sets expand, the runtime limitations of these traditional approaches have become more apparent. Therefore, accelerating CEL has been crucial in the field of machine learning, with diverse methods explored to overcome the computational costs inherent to this domain. 
Recent efforts have focused on addressing these challenges through neural and neuro-symbolic methods. Neural Class Expression Synthesis (NCES) \cite{kouagou2023neural}, represents a significant departure from traditional search-based methods. NCES employs neural networks to synthesize class expressions directly, avoiding the exhaustive refinement process and reducing computational overhead. NCES2 has further extended this approach \cite{kouagou2023neurala}, which broadens its applicability to more expressive description logics, such as $\mathcal{ALCHIQ}^\mathcal D$, enabling CEL for more complex and expressive ontologies. 

Another promising approach is Neuro-Symbolic Class Expression Learning (DRILL) \cite{demir2023neuro}, which addresses the limitations of heuristic-based methods like CELOE and OCEL. Traditional methods rely heavily on myopic heuristics and fixed rules for redundancy elimination and expression simplification, which can lead to suboptimal runtimes and memory inefficiencies. DRILL uses deep Q-learning to dynamically adapt the search process, reducing runtimes and memory requirements while enhancing scalability. This reinforcement learning paradigm overcomes the drawbacks of fixed heuristic rules, providing a more flexible and efficient framework for CEL.

Other noteworthy approaches to CEL include CLIP \cite{heist2019uncovering}, a pruning method that extends CELOE that integrates neural networks to predict an approximate solution length, thereby streamlining the concept learning process.  Similarly, DL-Focl \cite{rizzo2020class}, a variant of DL-Foil \cite{fanizzi2008dl}, introduces omission rates to prune the search space and further accelerate the learning process.

Despite recent progress, scaling CEL methods remains a major challenge due to persistent memory and runtime constraints \cite{demir2023neuro,kouagou2023neural}. Caching mechanisms offer a promising solution by optimizing concept retrieval—the primary computational bottleneck in CEL. 

\subsection{Cache Algorithms with Replacement Policies}

Caching has long been recognized as a powerful technique to avoid redundant computations by reusing precomputed objects. By storing frequently accessed or computationally expensive data for future use, caching systems significantly reduce computation time and enhance overall efficiency. This concept finds widespread applications in various domains, including storage systems, databases, and web servers \cite{megiddo2004outperforming}. 

Every caching mechanism requires dedicated memory space to store objects. However, memory resources are inherently limited, necessitating a strategy to manage the space effectively when the cache becomes full. Objects already present in the cache must be removed to accommodate new objects. This eviction process is governed by a replacement policy, which selects the objects to evict from the cache. This policy plays a crucial role in the efficiency of caching systems.
The choice of a replacement policy significantly impacts the effectiveness of a caching system. An optimal replacement policy minimizes computation time and resource usage, leading to improved overall system performance \cite{sethumurugan2021designing}. The theoretical ideal, known as Belady’s optimal replacement policy \cite{belady1966study}, evicts the object that will not be accessed for the longest period in the future. While Belady’s approach offers optimal performance, its practical implementation is infeasible because predicting future access is impossible in real-world scenarios.

Instead, practical cache replacement policies have been developed, each leveraging specific heuristics to make eviction decisions. Among these are policies based on frequency, where eviction decisions are determined by how often an object has been accessed. Notable examples include Least Frequently Used (LFU) and Most Frequently Used (MFU) strategies, along with their variants. LFU evicts objects with the lowest access frequency, while MFU does the opposite, removing the most frequently accessed objects. Another widely adopted category of policies is based on recency, which considers the timing of an object’s last access. For instance, the Least Recently Used (LRU) strategy evicts objects that have not been accessed for the longest time, while the Most Recently Used (MRU) strategy prioritizes removing the most recently accessed objects. Some other methods, like ARC (Adaptive Replacement Cache) \cite{megiddo2004outperforming}  have been proposed and it combines both recency and frequency policies. 

Policies based on the order of entrance into the cache form yet another category. These include First In, First Out (FIFO) and Last In, First Out (LIFO) strategies. FIFO removes the oldest object in the cache, whereas LIFO eliminates the most recently added object. Following the same idea as the ensemble learning technique in machine learning,  \cite{ari2002acme} used LRU, LFU and FIFO together to decide which objects should be evicted from the cache.


\subsection{Metrics}
The performance of a caching system is typically evaluated using the hit ratio and the total runtime with and without cache. The hit ratio is given by
\begin{equation}
\text{hit ratio} = \frac{H}{H + M}\, ,
\end{equation}
where $H$ and $M$ represent the number of cache hits (data found in the cache) and cache misses (data not found in the cache), respectively. The hit ratio measures the proportion of requests successfully served from the cache, indicating the efficiency of the caching strategy in reducing computational overhead. Conversely, the miss ratio ($1-\text{hit ratio}$) quantifies the proportion of requests that could not be served from the cache, reflecting the instances where the system had to compute or retrieve the required data anew. Cache algorithms aim to maximize hit ratios and consequently minimize the miss ratios, as they signify reduced reliance on expensive computations and improved system efficiency. 

Our work builds on these foundations by introducing a simple caching mechanism tailored specifically for semantic reasoning tasks. By focusing on the subsumption relationships between concepts, we provide a novel perspective on integrating caching into CEL pipelines. The next section delves deeper into our proposed algorithms with a focus on how they can be used to accelerate concept retrieval in CEL.

\section{Preliminaries}

\subsection{Description Logics and Knowledge Bases}

Description Logics (DLs), initially introduced by Brachman \cite{brachman1978structural}, are a family of knowledge representation languages that can be seen as decidable fragments of first-order logic \cite{lehmann2010concept}. They serve as a formalism for representing structured and well-understood knowledge \cite{nardi2003introduction,baader2009description}. DLs provide the foundation for modeling ontologies by enabling the expression of concepts (classes), roles (relationships), and individuals in a formal and logically grounded manner.

These logics range in expressivity, from simpler ones such as $\mathcal{EL}$ and $\mathcal{ALC}$ to more expressive variants like $\mathcal{SHOIQ}$ and $\mathcal{SROIQ}$, which underlie standards such as OWL DL.

A Description Logic-based knowledge base (KB) is typically defined as a pair $(\mathcal{T}, \mathcal{A})$, where $\mathcal{T}$ is the TBox (terminological box) and $\mathcal{A}$ is the ABox (assertional box). The TBox contains general concept inclusions, i.e., axioms of the form $C \sqsubseteq D$, capturing hierarchical relationships between concepts. The ABox, on the other hand, contains instance-level assertions such as $C(a)$ (individual $a$ is an instance of concept $C$) and $r(a,b)$ (individual $a$ is related to $b$ via role $r$), where $C, D$ are concepts, $r$ is a role, and $a, b$ are individuals.

\subsection{Class Expression Learning}

Let $N_I$ denote the set of individuals in a knowledge base $\mathcal K$. 
Given a description logic $\mathcal L$ (e.g., $\mathcal{ALC}$, $\mathcal{SHOIN}$, or $\mathcal{SROIQ}$) and two sets of labeled examples $E^+ \subseteq N_I$ (positive examples) and $E^- \subseteq N_I$ (negative examples), the goal of CEL is to construct a concept $H \in \mathcal C(\mathcal L)$ such that
\[
\forall e^+ \in E^+: \mathcal K \models H(e^+) 
\quad \text{and} \quad
\forall e^- \in E^-: \mathcal K \not\models H(e^-),
\]
where $\mathcal C(\mathcal L)$ denotes the set of all concepts that can be expressed in $\mathcal L$ \cite{kouagou2023neural,demir2023neuro,fanizzi2018dlfoil}. 
In practice, perfect separation is often impossible, and learners therefore aim to identify concepts that optimize a quality measure such as accuracy, precision, recall, or F1-score.

Most CEL algorithms explore the concept space $\mathcal C(\mathcal L)$ through heuristic search guided by refinement operators \cite{lehmann2010concept,lehmann2011class}. A common approach is top-down search, where the process starts from the most general concept (e.g., $\top$) and incrementally specializes it by applying refinement rules that introduce logical constructors such as conjunctions, disjunctions, or existential restrictions. Other approaches explore the space bottom-up or combine symbolic search with neural guidance \cite{fanizzi2018dlfoil,rizzo2020class,kouagou2023neural,heindorf2022evolearner}.

Regardless of the search strategy, evaluating candidate concepts is a central step in CEL. For each generated concept $C$, the learner must determine the set of individuals $C^I$ that satisfy the concept in $\mathcal K$. This operation corresponds to concept retrieval and typically requires invoking a description logic reasoner. The retrieved instances are then compared with $E^+$ and $E^-$ to compute the quality of the candidate concept. Prior work has identified concept retrieval as one of the main computational bottlenecks in CEL pipelines \cite{DBLP:conf/pkdd/KaralisBDHN24}. This observation motivates the development of mechanisms that can reuse previously computed reasoning results to reduce redundant calls to the reasoner.

 \paragraph{Notation:}
 Unless stated otherwise, we adopt the following notations.  Given a concept $C$, $Ret(C)$ denotes the retrieved instances of $C$. Given a set $\mathcal{S}$, $|\mathcal{S}|$ is the cardinal of $\mathcal{S}$. $N_I$, $N_C$, and $N_R$ represent the set of named individuals, concepts, and roles, respectively. 
 
 \section{Approach}

Our goal is to accelerate concept retrieval in description logic reasoners by introducing a semantics-aware cache. While the approach can in principle be extended to more expressive logics, we focus on $\mathcal{ALC}$ in this work because it provides the core constructors required by many CEL systems while keeping the exposition clear. The syntax and semantics of $\mathcal{ALC}$ concepts are summarized in Table~\ref{tab:semantics1}. 

\begin{table}[htb]
	\centering
    \caption{Syntax \& semantics for $\mathcal{ALC}$ concepts. 
$\mathcal{I}$ stands for an interpretation with domain $\Delta^\mathcal{I}$. 
 }  
    
   \begin{tabular}{@{}lccc@{}}
        \toprule
		\textbf{Construct}               & \textbf{Syntax}         & \textbf{Semantics} \\
		\midrule
        Atomic concept          & $A$            & $A^{\mathcal{I}}\subseteq{\Delta^\mathcal{I}}$ \\
        Atomic role                    & $r$            & $r^\mathcal{I}\subseteq{\Delta^\mathcal{I}\times \Delta^\mathcal{I}}$\\
  		Top concept             & $\top$         & $\Delta^\mathcal{I}$\\
  		Bottom concept          & $\bot$         & $\emptyset$            \\
 		
 		Negation                & $\neg C$       & $\Delta^\mathcal{I}\setminus C^\mathcal{I}$
 		\\
 		Conjunction             & $C\sqcap D$    & $C^\mathcal{I}\cap D^\mathcal{I}$ \\
 		Disjunction             & $C\sqcup D$    & $C^\mathcal{I}\cup D^\mathcal{I}$\\
 		Existential restriction & $\exists\  r.C$ & $\{ x \mid \exists\ y. (x,y) \in 
   r^\mathcal{I} \land y \in C^\mathcal{I}\}$\\
 		Universal restriction & $\forall\   r.C$   & $\{ x \mid \forall\  y. (x,y) \in r^\mathcal{I} \implies y \in C^\mathcal{I}\} $\\
		\bottomrule
	\end{tabular}

 \label{tab:semantics1}
\end{table}

\subsection{Overview}

Our approach exploits a fundamental property of description logics: if a concept $C$ is subsumed by another concept $D$, then every instance of $C$ must also be an instance of $D$, i.e.,

\[
C \sqsubseteq D \Rightarrow Ret(C) \subseteq Ret(D).
\]

A naive concept retrieval procedure evaluates $C(a)$ for every individual $a \in N_I$ to compute $Ret(C)$. However, if we already know the set $Ret(D)$ for a concept $D$ with $C \sqsubseteq D$, we only need to verify the membership of individuals in $Ret(D)$ to determine $Ret(C)$. When $|Ret(D)| \ll |N_I|$, this can significantly reduce the number of reasoning calls.

This observation motivates our caching strategy. Instead of recomputing concept retrieval queries from scratch, the cache stores previously evaluated concepts together with their instance sets. When a new query concept $C$ is issued, the cache attempts to identify previously stored concepts that subsume $C$ and uses their instance sets to restrict the search space.

The main challenge lies in identifying such concepts efficiently. Determining whether $C \sqsubseteq D$ is NP-hard in $\mathcal{ALC}$ \cite{ceylan2017bayesian}. Performing general subsumption checks during cache lookup would therefore defeat the purpose of the cache. Instead, we employ a small set of sound heuristics that detect subsumption relationships through simple syntactic patterns. These heuristics operate only on concepts already stored in the cache and avoid invoking a reasoner.

 \subsection{Subsumption Heuristics and Correctness}

The heuristics used by our cache are designed to identify subsumption relations through simple structural patterns in concept expressions. Importantly, whenever a heuristic concludes that $C \sqsubseteq D$, the subsumption relation is guaranteed to hold under the standard semantics of description logics.

The heuristics used in our implementation are the following:

\begin{itemize}

\item If $C \equiv D \sqcap E$, then $C \sqsubseteq D$.

\item If $C \equiv \exists r.C'$ and $D \equiv \exists r.\top$, then $C \sqsubseteq D$.

\item If $C \sqsubseteq D_1, \ldots, C \sqsubseteq D_n$, then  $C \sqsubseteq \bigsqcap_{i=1}^n D_i .$ 

\end{itemize}

These rules follow directly from the semantics of $\mathcal{ALC}$. For instance, the first rule holds because the interpretation of a conjunction is defined as the intersection of the interpretations of its conjuncts. Similarly, the existential rule follows from the fact that $\exists r.C'$ always implies $\exists r.\top$.

During cache lookup, these heuristics are applied to concepts already present in the cache. The cache is implemented as a hash map whose keys correspond to the abstract syntax trees of concept expressions, see Algorithm \ref{algo:semantic}. The heuristics are evaluated over these keys to identify potential superconcepts of the queried concept.

Importantly, the system does not perform general subsumption reasoning. All checks are local and syntactic, ensuring that cache lookup remains computationally lightweight while preserving correctness. 

Once candidate superconcepts $D_i$ have been identified in the cache, their instance sets are intersected to obtain a reduced candidate set:

\[
S = \bigcap_i Ret(D_i).
\]

Instead of evaluating $C(a)$ for all $a \in N_I$, the reasoner only needs to check individuals in $S$. If an exact match $C \equiv D$ is found among the cache keys, the cache can directly return $Ret(D)$ without invoking the reasoner.

This recursive decomposition, combined with subsumption-based pruning, yields a constructive correctness argument: every individual returned by the cache satisfies the semantics of the queried concept, while individuals outside the candidate set are guaranteed not to belong to the result.
 
\subsection{Cache Initialization}

Although the semantic cache can operate without an initialization phase, pre-populating the cache with frequently used concepts significantly improves its effectiveness. In particular, CEL systems often generate candidate expressions built from atomic concepts and simple role restrictions. Precomputing these instance sets, therefore, increases the probability that subsequent queries can reuse cached results.

When initialization is performed, we add the following precomputed concept retrieval results to the cache: 
\begin{itemize}
	\item $Ret(A)$ for all $A \in N_C$
        \item $Ret(\neg A)$ for all $A \in N_C$
	\item $Ret(\exists r.C)$ for all $r \in N_R$ and $C \in N_C \cup \{\top\}$
\end{itemize}
The details of this implementation is shown in Algorithm \ref{algo:init}.

\begin{algorithm}
	\caption{initalizeCache()}
\begin{algorithmic}[1]
    \ForAll{$r \in N_R$}
        \State store($\exists r.\top$, Ret($\exists r.\top$))
    \EndFor
    \ForAll{$A \in N_C$}
		\State store($A$, Ret($A$))
            \State store($\neg A$, $N_I\setminus Ret(A)$)
		\ForAll{$r \in N_R$}
		\State store($\exists r.A$, Ret($\exists r.A$))
		\EndFor
	\EndFor
\end{algorithmic}
 \label{algo:init}
\end{algorithm} 

\subsection{Fetching from Cache and Space Management}

Algorithms~\ref{algo:semantic} and~\ref{algo:store} describe how concept retrieval queries are answered using the cache and how cache entries are managed. The procedure \texttt{fetchInstances} first checks whether the queried concept $C$ is already stored in the cache. For named concepts, the corresponding instance sets are typically precomputed during initialization and can therefore be returned immediately.

If the concept is not directly available, the algorithm evaluates it by recursively decomposing the expression according to the semantics of $\mathcal{ALC}$. Negations are handled by computing the complement of the retrieved instances, conjunctions and disjunctions correspond to set intersection and union, respectively, and existential restrictions are evaluated by checking for role successors that satisfy the filler concept. Universal restrictions are reduced to negated existential restrictions, i.e., $\forall r.C \equiv \neg(\exists r.\neg C)$.

If the result for $C$ is not already stored in the cache, the reasoner is invoked to compute its instance set, and the resulting pair $(C, Ret(C))$ is inserted into the cache. This ensures that subsequent queries involving the same concept can reuse previously computed results.

Cache capacity is controlled by Algorithm~\ref{algo:store}. When inserting a new entry, the algorithm first checks whether sufficient space is available. If the maximum capacity would be exceeded, a purge operation removes entries according to the selected eviction policy. The new concept and its instance set are then stored in the cache.

\begin{algorithm}
	\caption{store (Concept $C$, Set $R$)}\
	\begin{algorithmic}[1]
		\If{$R.size()+this.size() > maxSize$}
			\State purge($maxSize-R.size()$)
		\EndIf
			\State add($C$, $R$)
		
	\end{algorithmic}
 \label{algo:store}
\end{algorithm}

\begin{algorithm}[tb]
	\caption{Semantics-based implementation of fetchInstances(Concept $C$)}\
	\begin{algorithmic}[1]
		\If{$C \notin N_C$}
			\If{$C \equiv \top$}
				\State \Return $N_I$
			\ElsIf{$C \equiv \bot$}
			\State \Return $\emptyset$
			\ElsIf{$C \equiv \neg D$}
			 \State \Return  $N_I \setminus$ fetchInstances($D$)
			\ElsIf{$C \equiv D \sqcap E$}
				\State \Return fetchInstances($D$) $\cap$ fetchInstances($E$)
			\ElsIf{$C \equiv D \sqcup E$}
				\State \Return fetchInstances($D$) $\cup$ fetchInstances($E$)
			\ElsIf{$C \equiv \exists r.D$}
				\State Set instances = fetchInstances($D$)
				\State Set $R = \emptyset$
				\For{a $\in N_I$ }
					\For{b $\in$ instances}
						\If{reasoner.check($r(a,b)$)}
							\State $R$.add($a$)
							\State break
						\EndIf
					\EndFor
				\EndFor
				\State \Return R
			\ElsIf{$C \equiv \forall r.D$}
			\State \Return fetchInstances($\neg (\exists r. \neg D$))
		\EndIf
		\EndIf
		\If{$\neg$ keys.contains($C$)}
		\State Set $R$ = reasoner.getInstances($C$)
		\State store($C$, $R$)
		\EndIf 
		\State return  get($C$)
	\end{algorithmic}
 \label{algo:semantic}
\end{algorithm}


\section{Experiments}

\paragraph{Tasks and Baselines}
Our experiments evaluate the impact of our caching system on concept retrieval and concept learning tasks. We compare three configurations: (i) no cache, (ii) our semantic cache, and (iii) a non-semantic cache. The latter uses the same storage interface but does not exploit the syntactic or semantic structure of concept expressions, effectively corresponding to naïve memoization. To the best of our knowledge, no caching mechanism has been specifically proposed for class expression learning or concept retrieval. We therefore use these three configurations as baselines to assess both the benefits of caching and the additional gains from semantic reasoning.

For concept retrieval, concepts with a maximum length of 10 are generated by randomly sampling concepts from the refinement tree produced by any concept learner. The total number of generated concepts per dataset is summarized in Table \ref{tab:datasets}. For concept learning, we rely on existing learning problems (LPs), which are provided in the supplementary material.

We evaluate our approach using several state-of-the-art reasoners commonly used in concept retrieval frameworks. The symbolic reasoners include JFact \cite{tsarkov2006fact++}, HermiT \cite{glimm2014hermit}, Pellet \cite{sirin2007pellet}, and its extended version Openllet \cite{singh2020owl2bench}. In addition, we include a neural reasoning approach (\approach) \cite{teyou2025neural} to investigate the effectiveness of the cache in a neuro-symbolic setting.

Our evaluation proceeds in two stages. First, we measure the performance of the reasoners without caching to establish a baseline. We then enable the cache and vary its size to analyze its impact on runtime and retrieval efficiency. The cache size is defined relative to the number of concepts generated for a dataset. For instance, a cache size of $10\%$ indicates that the cache can store up to $10\%$ of the total number of generated concepts (see Table~\ref{tab:datasets}). In addition to runtime performance, we also measure the physical memory consumption of the cache for both the semantic and non-semantic variants in order to quantify the average memory requirements of the caching mechanism.

We also analyze the impact of cache initialization. On the small Family dataset, we evaluate the cache both with and without the initialization phase to measure its contribution. For larger datasets, however, the cache is used only with initialization enabled, as this configuration consistently yields better performance and more stable behavior in practice.

\paragraph{Hardware} All experiments were conducted on a virtual machine equipped with 2 NVIDIA H100-80C GPUs (80GB each) and 32 Intel Xeon Platinum CPUs with 31GB of memory. The GPUs are required for batch embedding computation in the neuro-symbolic reasoner \approach.

\paragraph{Datasets}
We evaluate our cache on four benchmark datasets: Family, Carcinogenesis, Mutagenesis, and Vicodi \cite{bin2016towards}. Family is relatively small and commonly used for testing algorithms involving family relationships. To evaluate scalability, we also use larger datasets: Vicodi, Carcinogenesis, and Mutagenesis, which are widely used in bioinformatics and cheminformatics and contain detailed information about chemical compounds and their biological properties \cite{kouagou2023neurala}. Key dataset statistics are reported in Table~\ref{tab:datasets}.

\begin{table}[tb]
\centering
\caption{Datasets statistics. $|\textbf{NC}|$ is the total number of concepts generated on each dataset. $|$\textbf{Ind.}$|$ is the total number of individuals and  $|$\textbf{Prop.}$|$ is the total number of properties (relations) in each dataset.}
\setlength{\tabcolsep}{2.5pt}
\begin{tabular}{@{}lccccccc@{}}
    \toprule
    \textbf{Dataset} & $|$\textbf{Ind.}$|$ & $|$\textbf{Classes}$|$ & $|$\textbf{Prop.}$|$ & $|\textbf{TBox}|$ & $|\textbf{ABox}|$& $|\textbf{NC}|$\\
    \midrule
    Vicodi & 33,238 & 194 & 10 & 204 & 116,181 & 2,105  \\
    Carcinogenesis & 22,372 & 142 &\phantom{0}4 & 144 &\phantom{0}74,223 & 1,684 \\
    Mutagenesis & 14,145 &\phantom{0}86 &\phantom{0}5 & \phantom{0}82 &\phantom{0}47,722 & 1,572\\
    Family & \phantom{00}202 & \phantom{0}18 & \phantom{0}4 & \phantom{0}26 & \phantom{000}472&1,236 
    \\
    \bottomrule
\end{tabular}

\label{tab:datasets}
\end{table}

\begin{figure}[htb]
    \centering  \subcaptionbox*{Carcinogenesis}
{\includegraphics[width=0.35\textwidth]{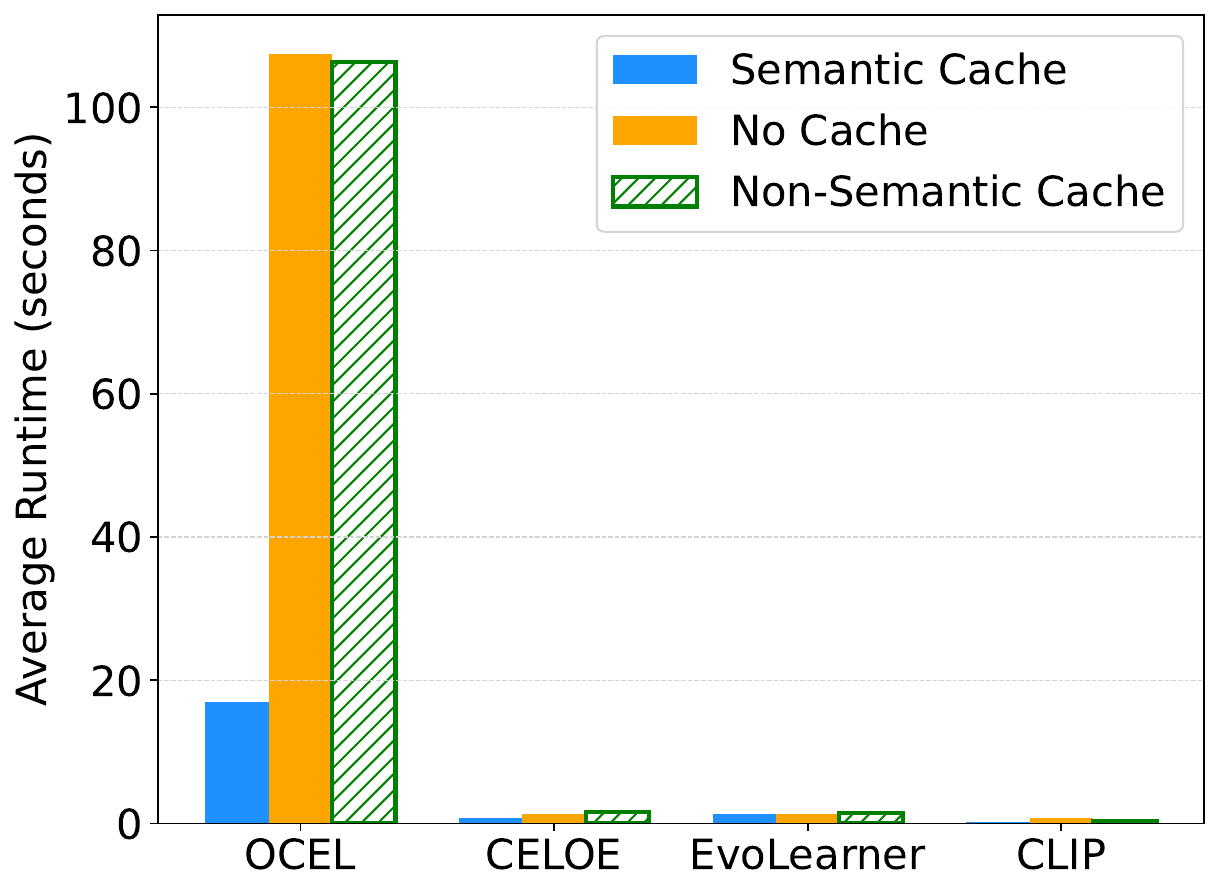}}
\subcaptionbox*{Vicodi}{\includegraphics[width=0.35\textwidth]{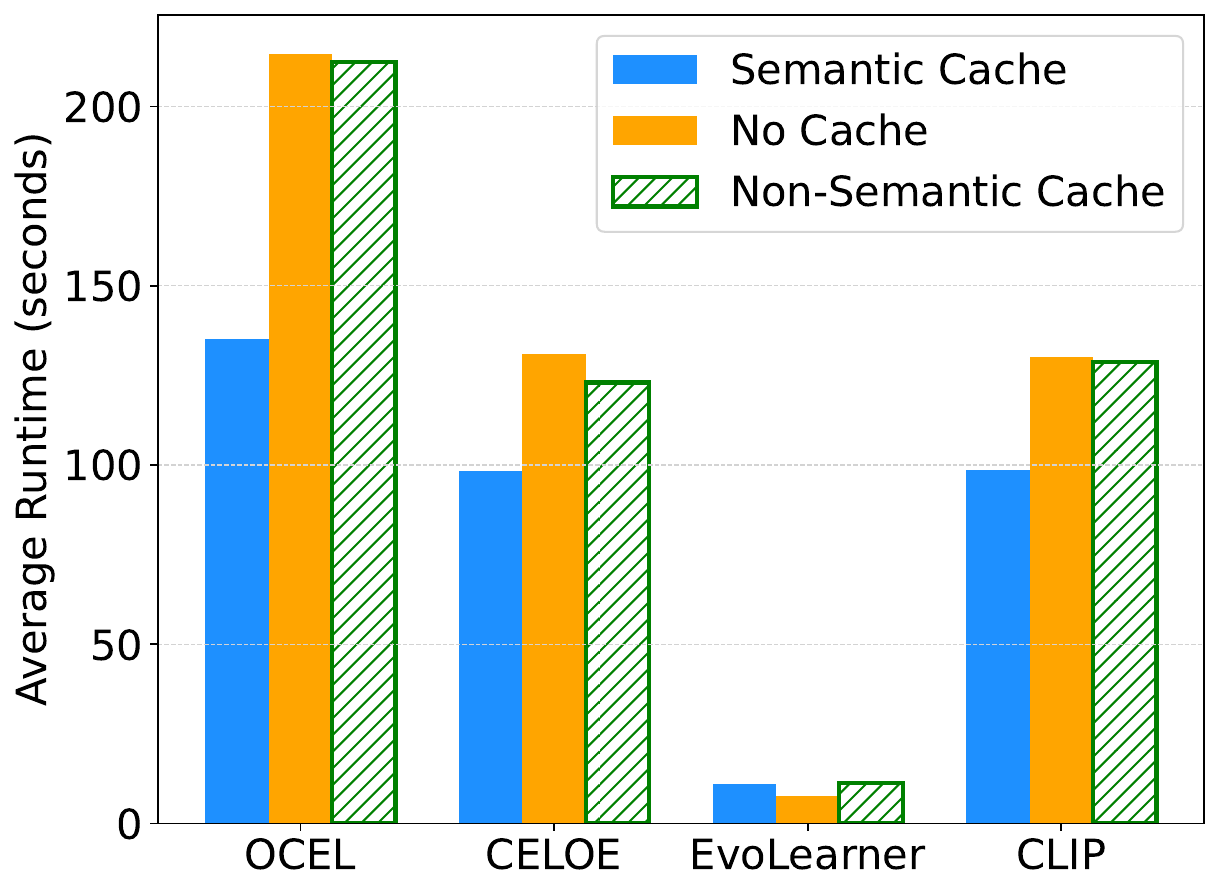}}
\subcaptionbox*{Mutagenesis}{\includegraphics[width=0.35\textwidth]{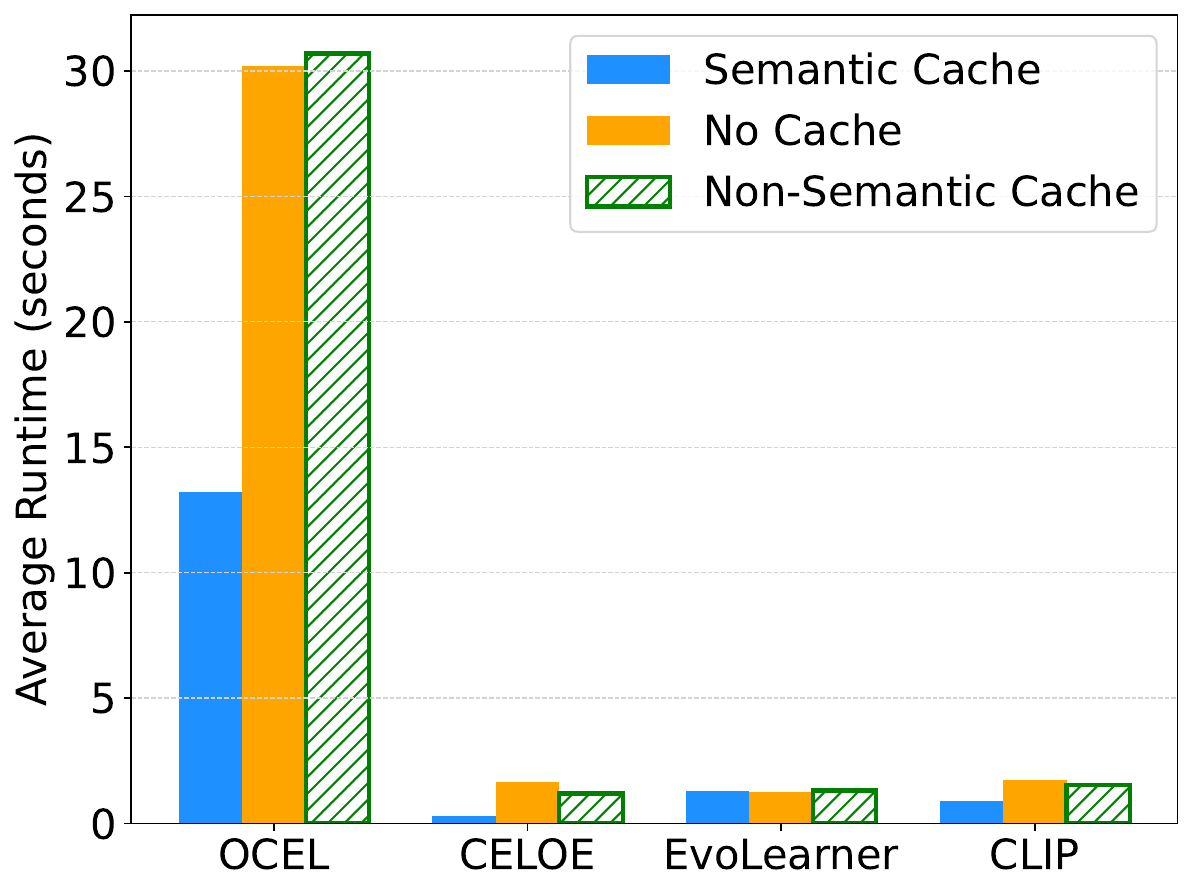}}
\subcaptionbox*{Family}{\includegraphics[width=0.35\textwidth]{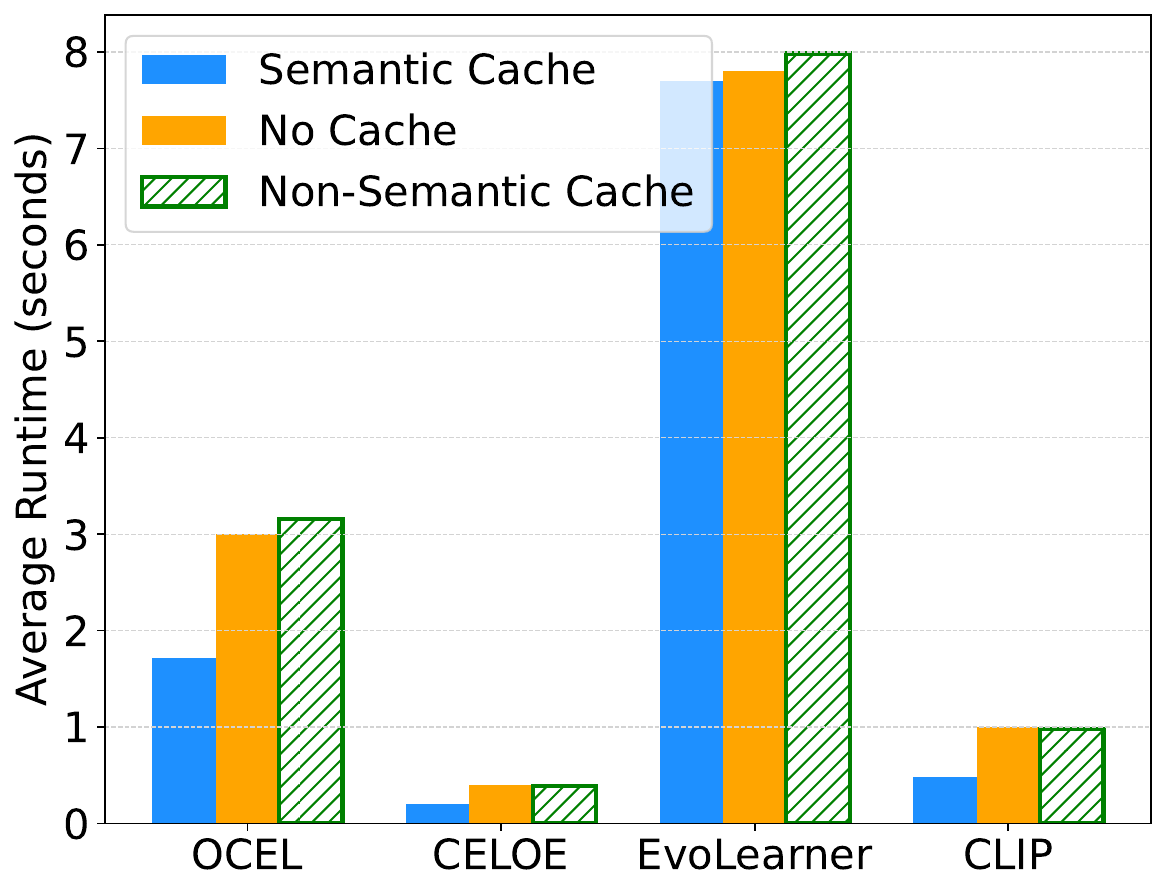}}
 \caption{Average runtime of concept learners using no cache, the semantic cache, and the non-semantic cache baseline across different datasets.}
    \label{fig:cel_results_cache}
\end{figure}

\section{Results and Discussion}
The results aim to demonstrate the impact of integrating our caching system on the performance of reasoners across various datasets. The evaluation focuses on two critical metrics: runtime and hit ratio as functions of cache size. The results section is structured into three parts. First, we integrate our caching system with five distinct eviction strategies—FIFO, LIFO, LRU, MRU, and RP—on small datasets, highlighting the differences in reasoner speed with and without caching. This analysis illustrates how the choice of cache policy interacts with our caching algorithm. Next, we identify the best-performing eviction strategy and demonstrate its effectiveness in enhancing reasoner performance on larger datasets when combined with our caching algorithm. Finally, we show how the cache can be easily applied to speed up concept learning algorithms.

\subsection{Results on Concept Retrieval}

\paragraph{Runtime analysis}

\begin{figure*}[htb]
    \centering
   \begin{minipage}[t]{0.44\textwidth}
        \centering
    \subcaptionbox*{}{\includegraphics[width=0.49\textwidth]{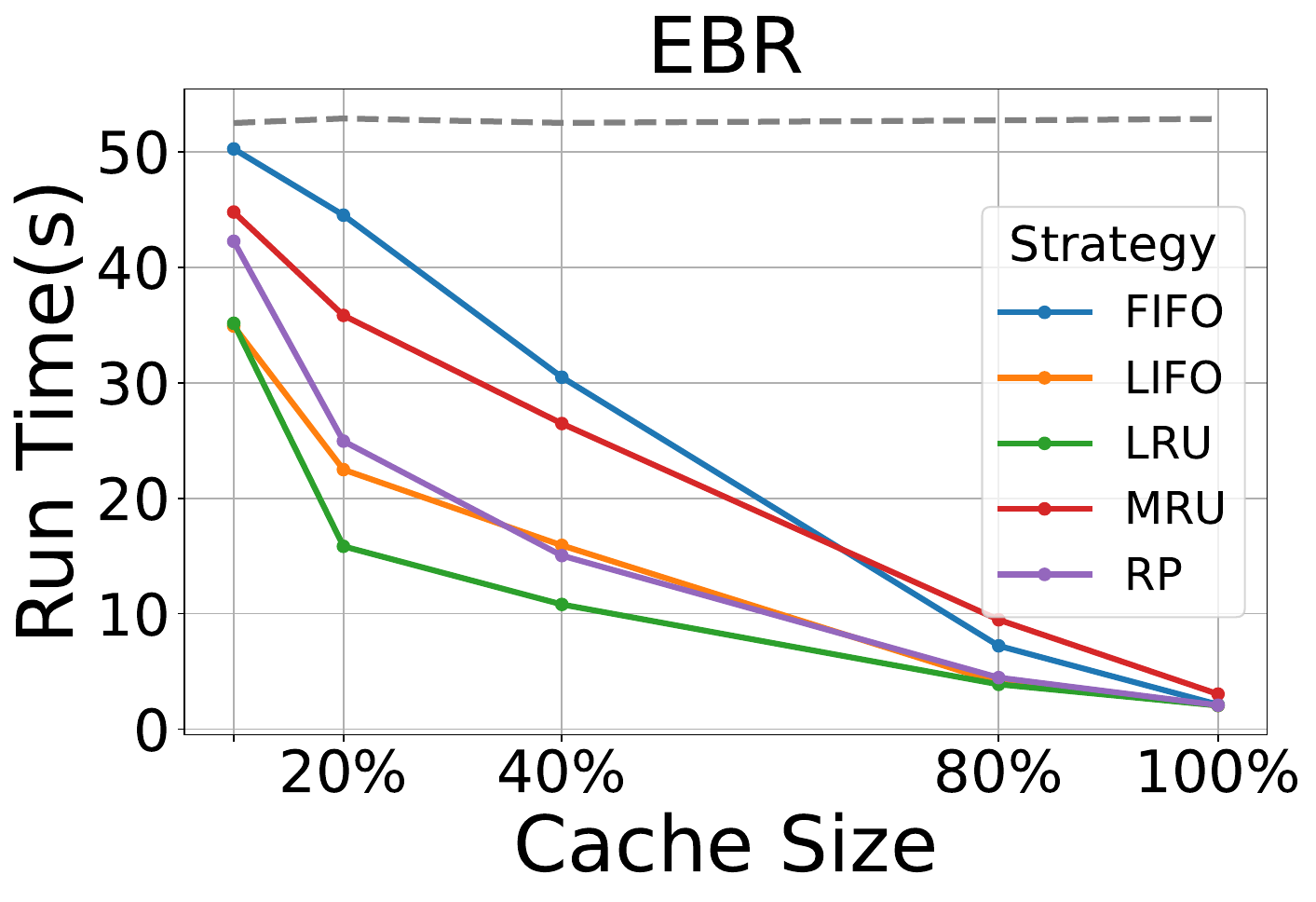}}
    \subcaptionbox*{}{\includegraphics[width=0.49\textwidth]{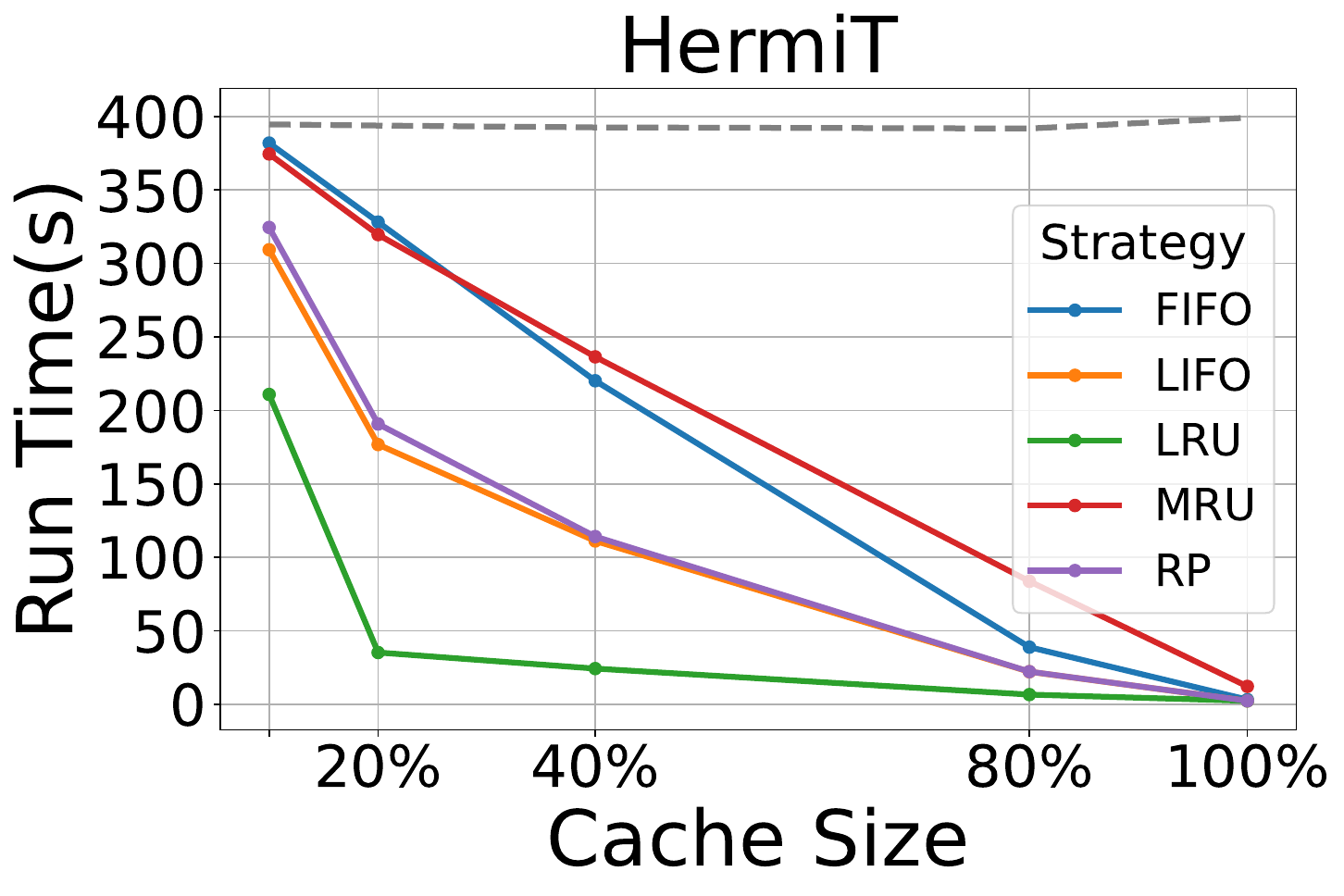}}
    \subcaptionbox*{}{\includegraphics[width=0.49\textwidth]{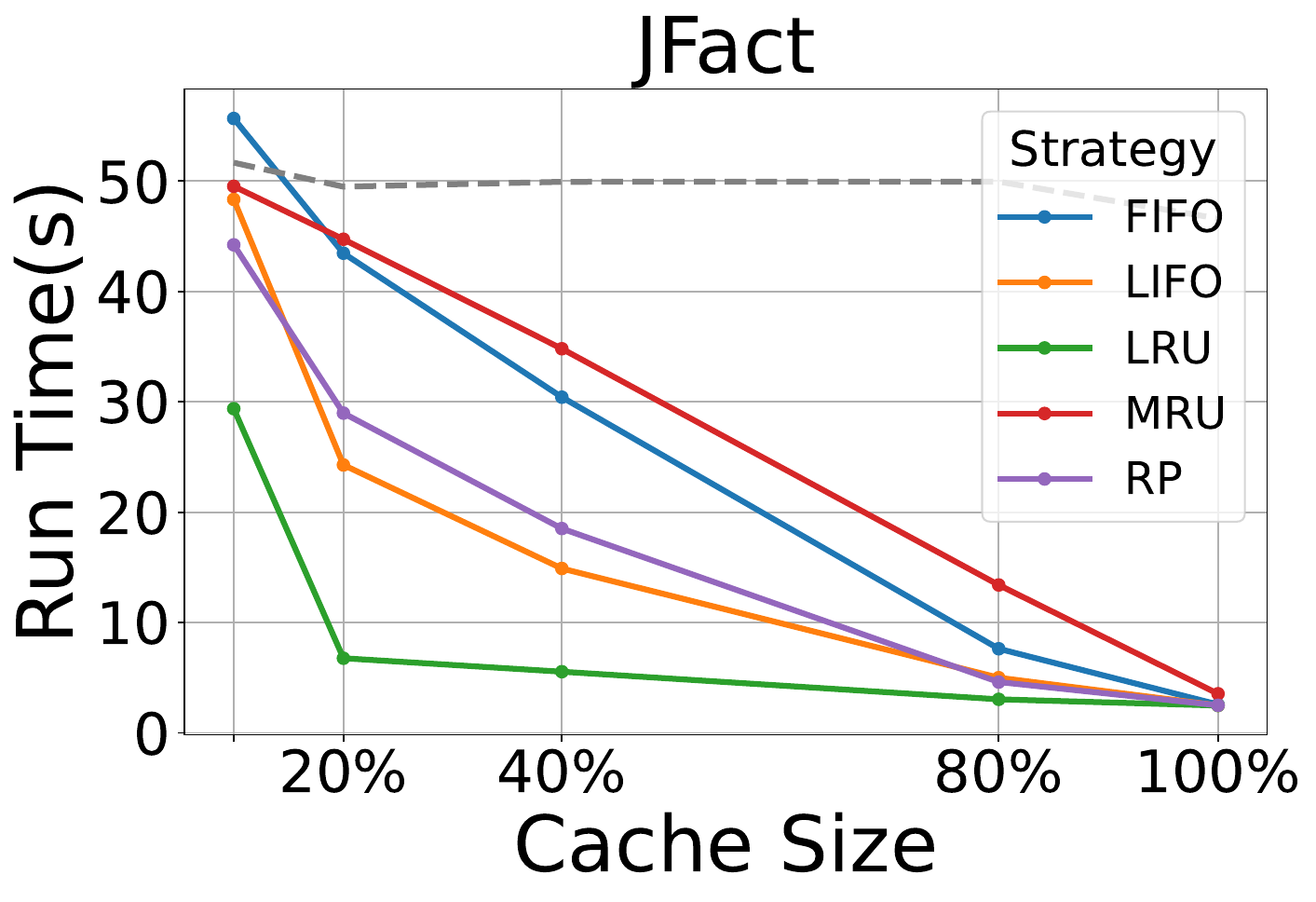}}
    \subcaptionbox*{}{\includegraphics[width=0.49\textwidth]{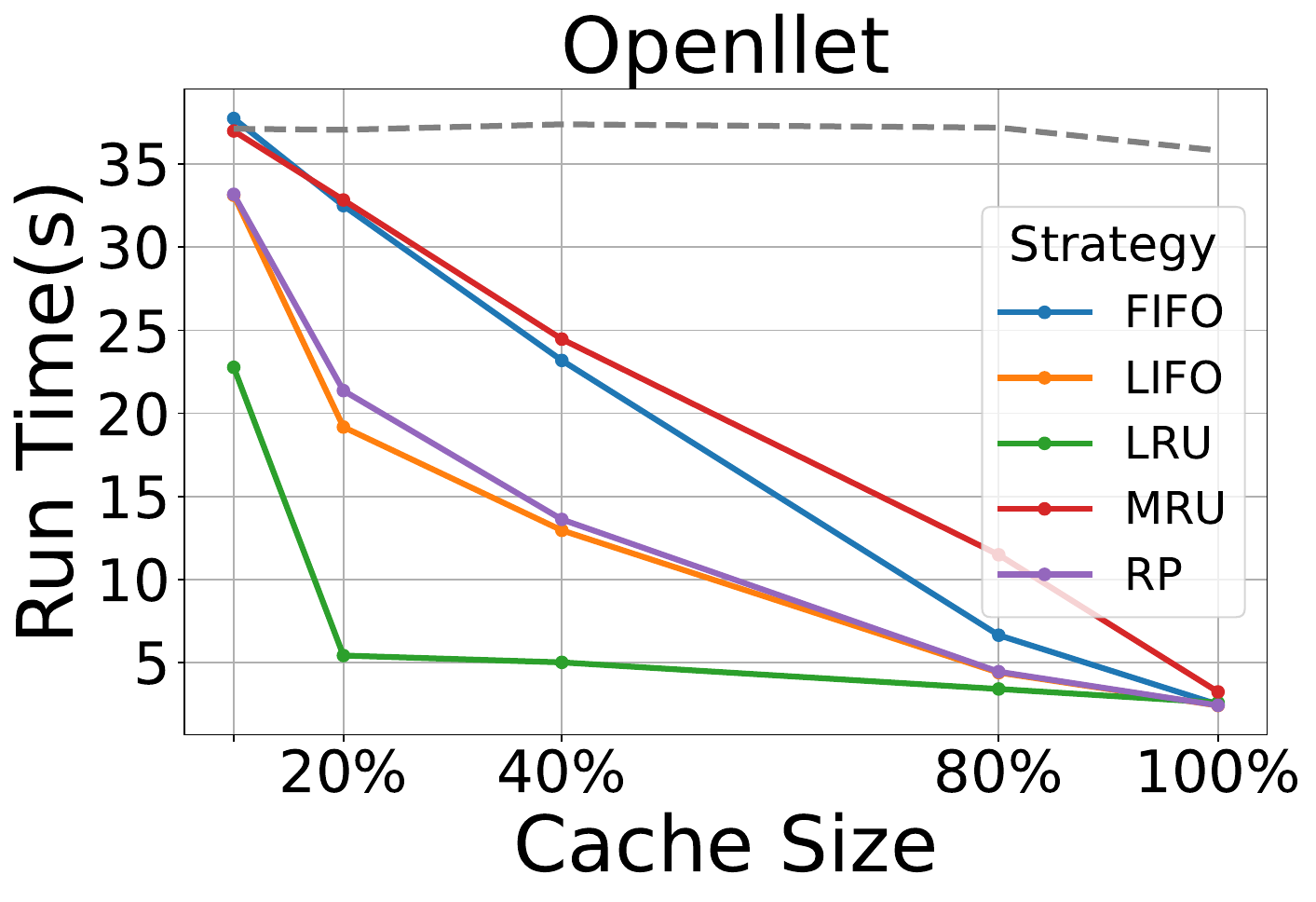}}
    \caption{Run time performance vs. cache size for each reasoner on the Family dataset with cache initialization. The gray dotted line indicates the runtime of the reasoner without our cache. }
    \label{fig:result_RT_family_init}
    \end{minipage}
     \hfill
     \begin{minipage}[t]{0.44\textwidth}
        \centering
        \subcaptionbox*{}{\includegraphics[width=0.49\textwidth]{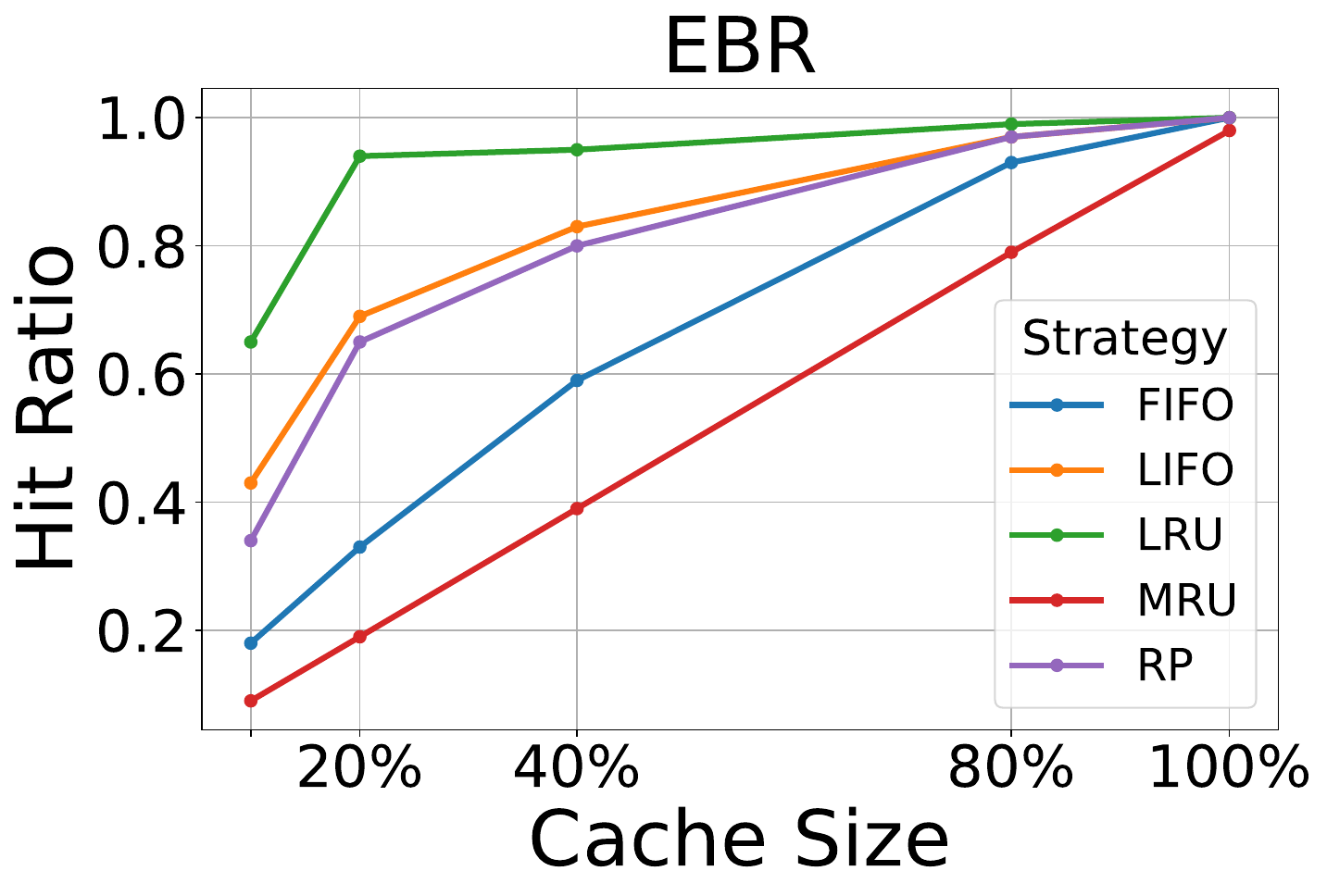}}
        \hfill
        \subcaptionbox*{}{\includegraphics[width=0.49\textwidth]{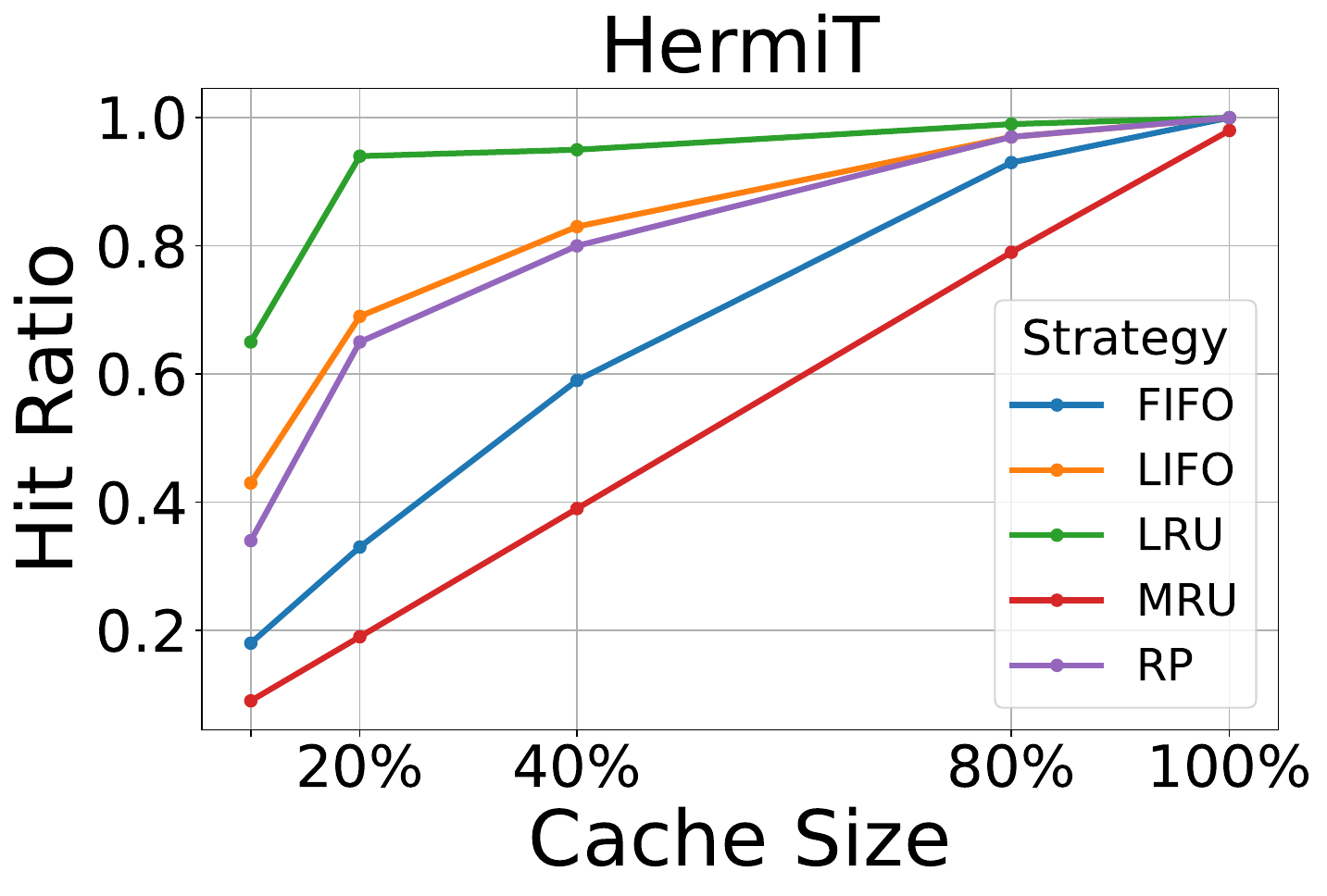}}
        
        \subcaptionbox*{}{\includegraphics[width=0.49\textwidth]{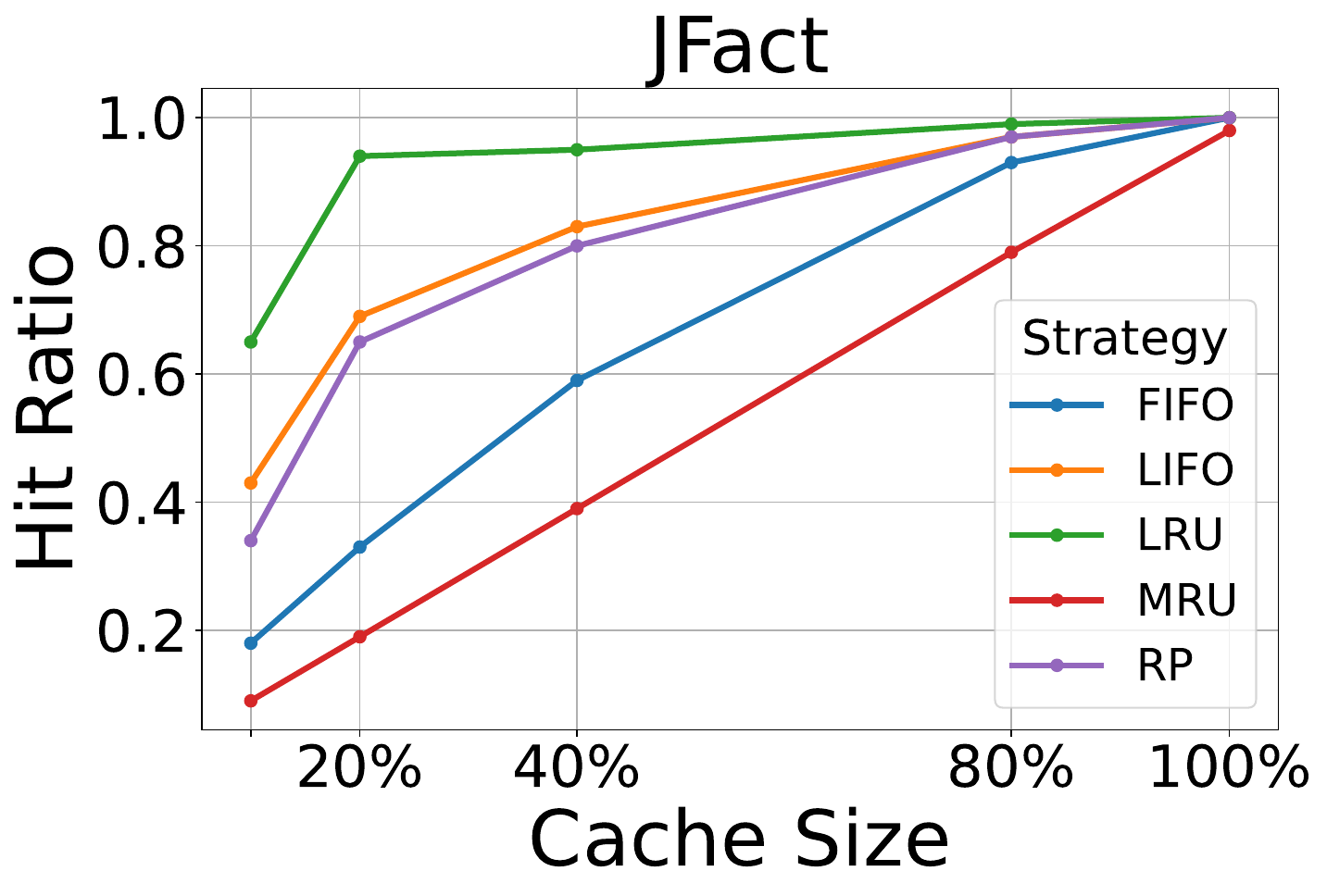}}
        \hfill
        \subcaptionbox*{}{\includegraphics[width=0.49\textwidth]{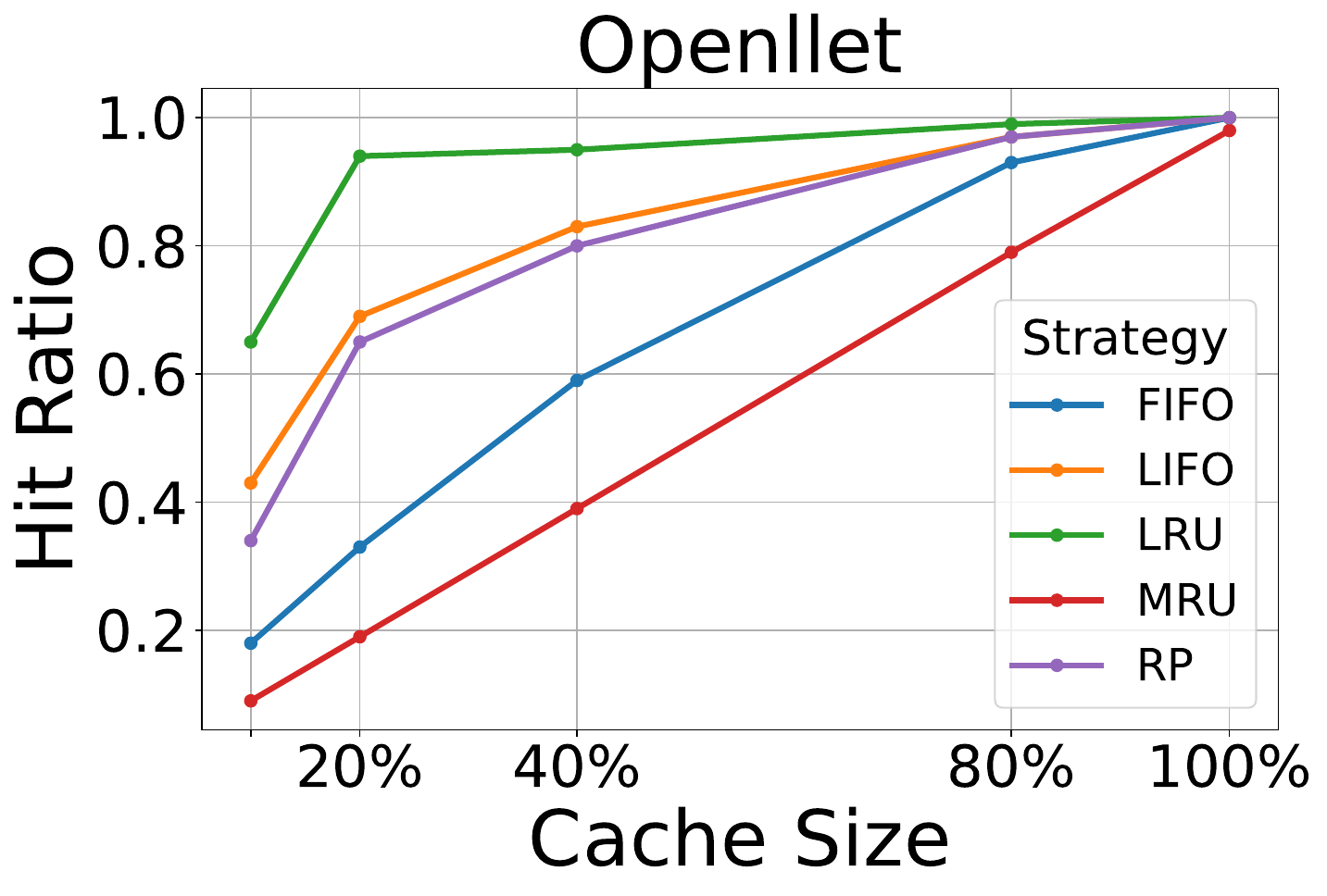}}
        
        \caption{Hit score performance vs cache size for each reasoner on the Family dataset with cache initialization. Higher values indicate more cache hits during retrieval.}
        \label{fig:result_hit_family_init}
    \end{minipage}
   
\end{figure*}

\begin{figure*}[htb]
    \centering
   \begin{minipage}[t]{0.44\textwidth}
        \centering
    \subcaptionbox*{}{\includegraphics[width=0.49\textwidth]{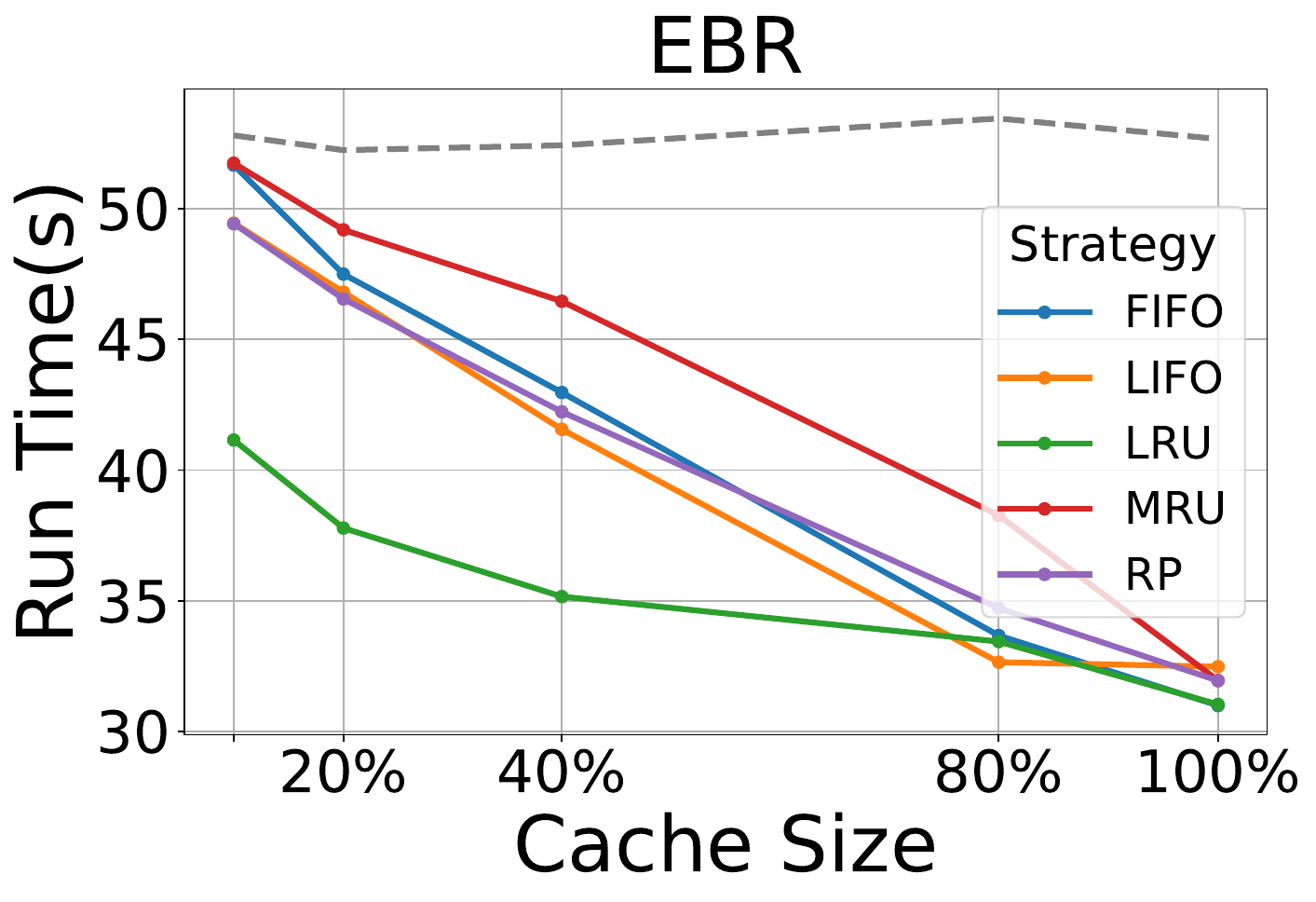}}
    \subcaptionbox*{}{\includegraphics[width=0.49\textwidth]{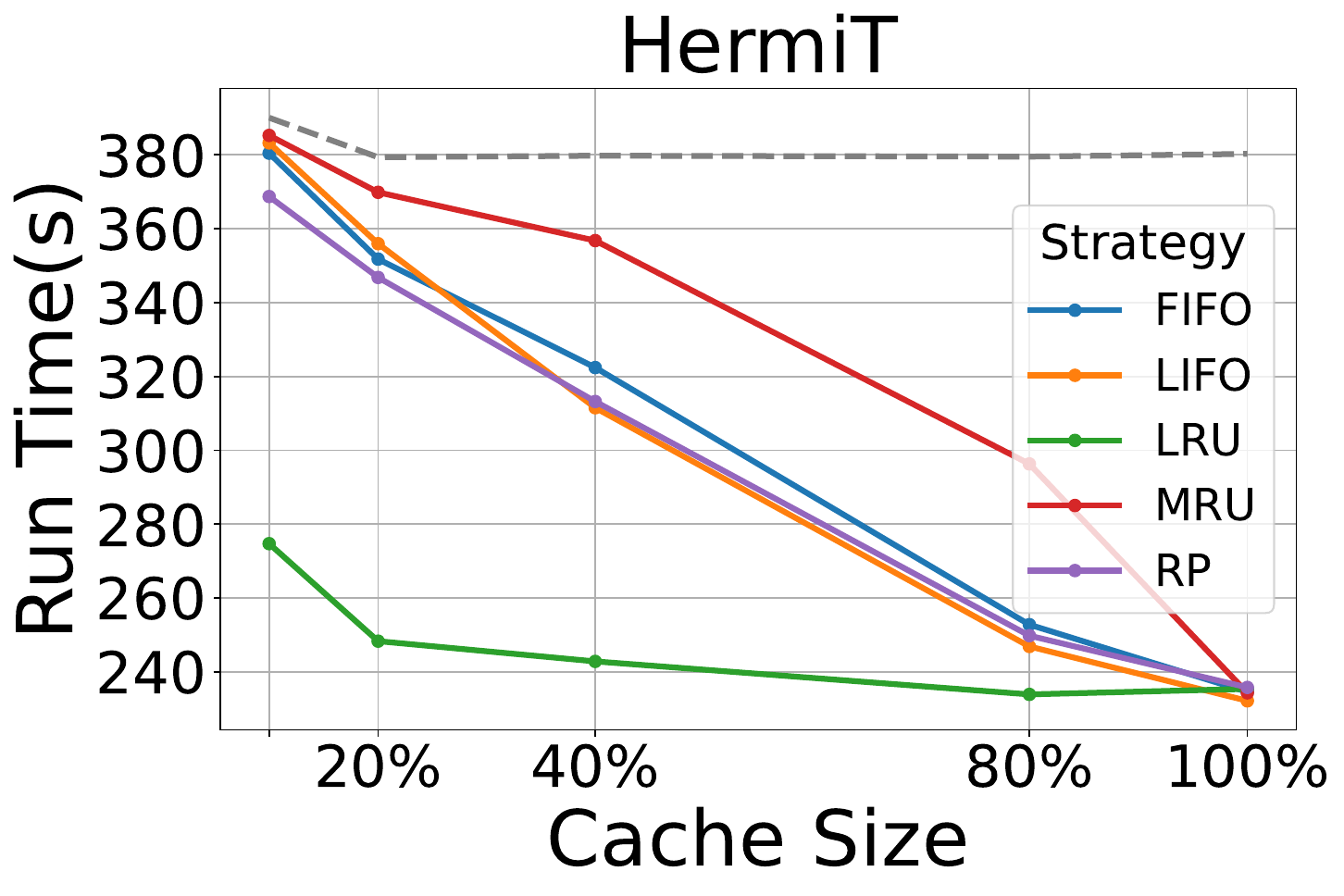}}
    \subcaptionbox*{}{\includegraphics[width=0.49\textwidth]{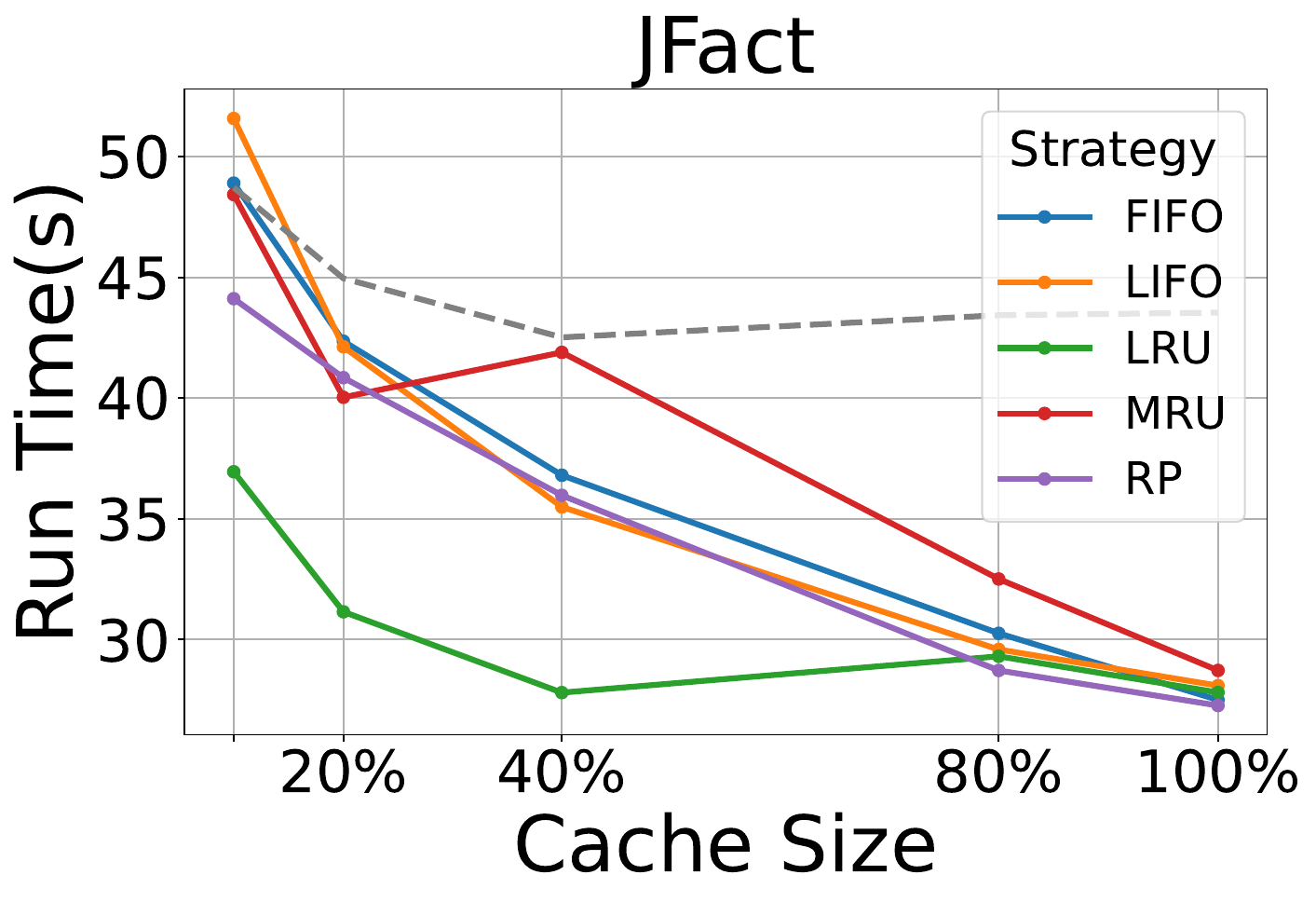}}
    \subcaptionbox*{}{\includegraphics[width=0.49\textwidth]{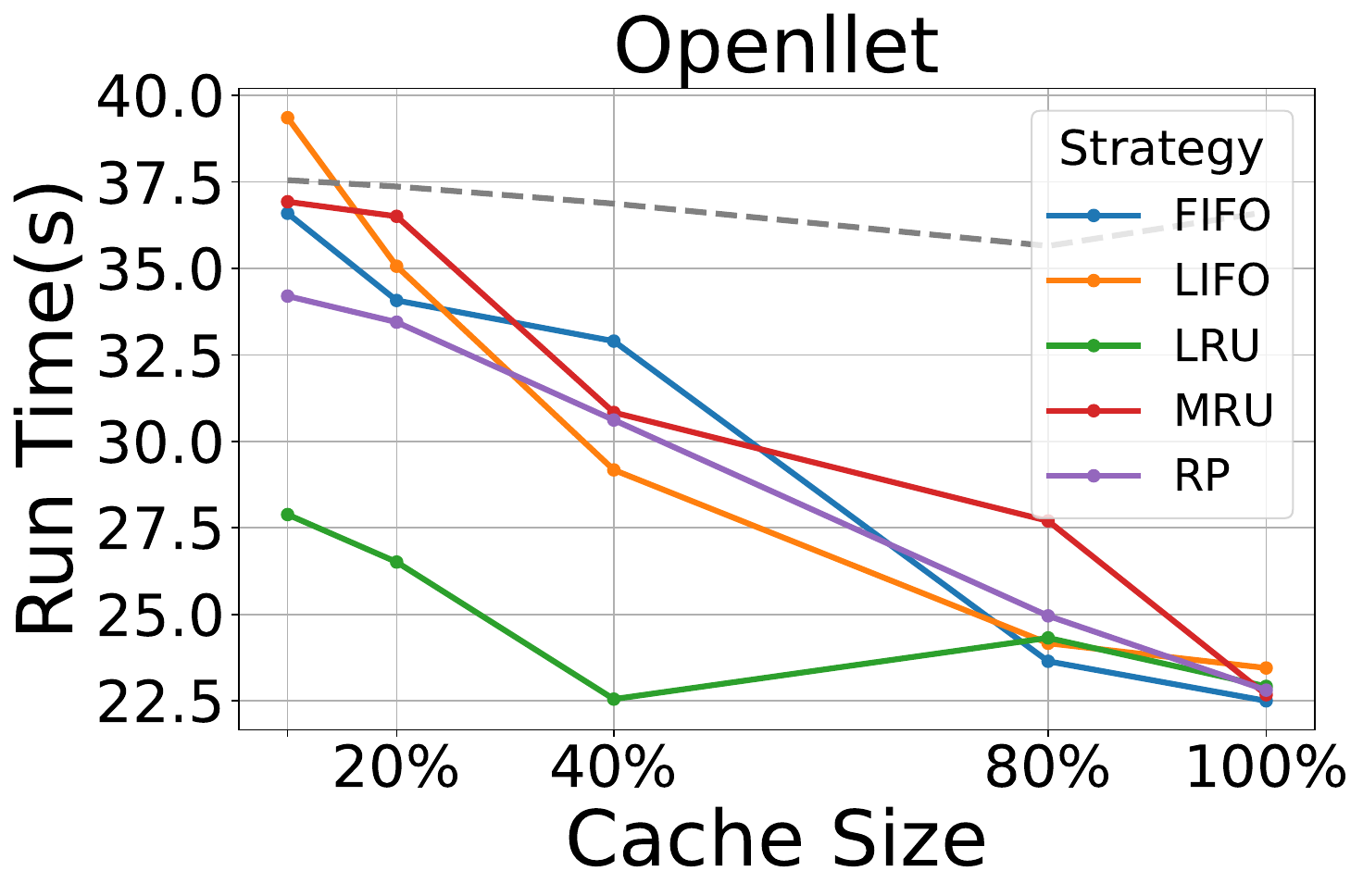}}
    \caption{Run time performance vs. cache size for each reasoner on the Family dataset without cache initialization. The gray dotted line indicates the runtime of the reasoner without our cache. }
    \label{fig:result_RT_family_no_init}
    \end{minipage}
     \hfill
     \begin{minipage}[t]{0.44\textwidth}
        \centering
        \subcaptionbox*{}{\includegraphics[width=0.49\textwidth]{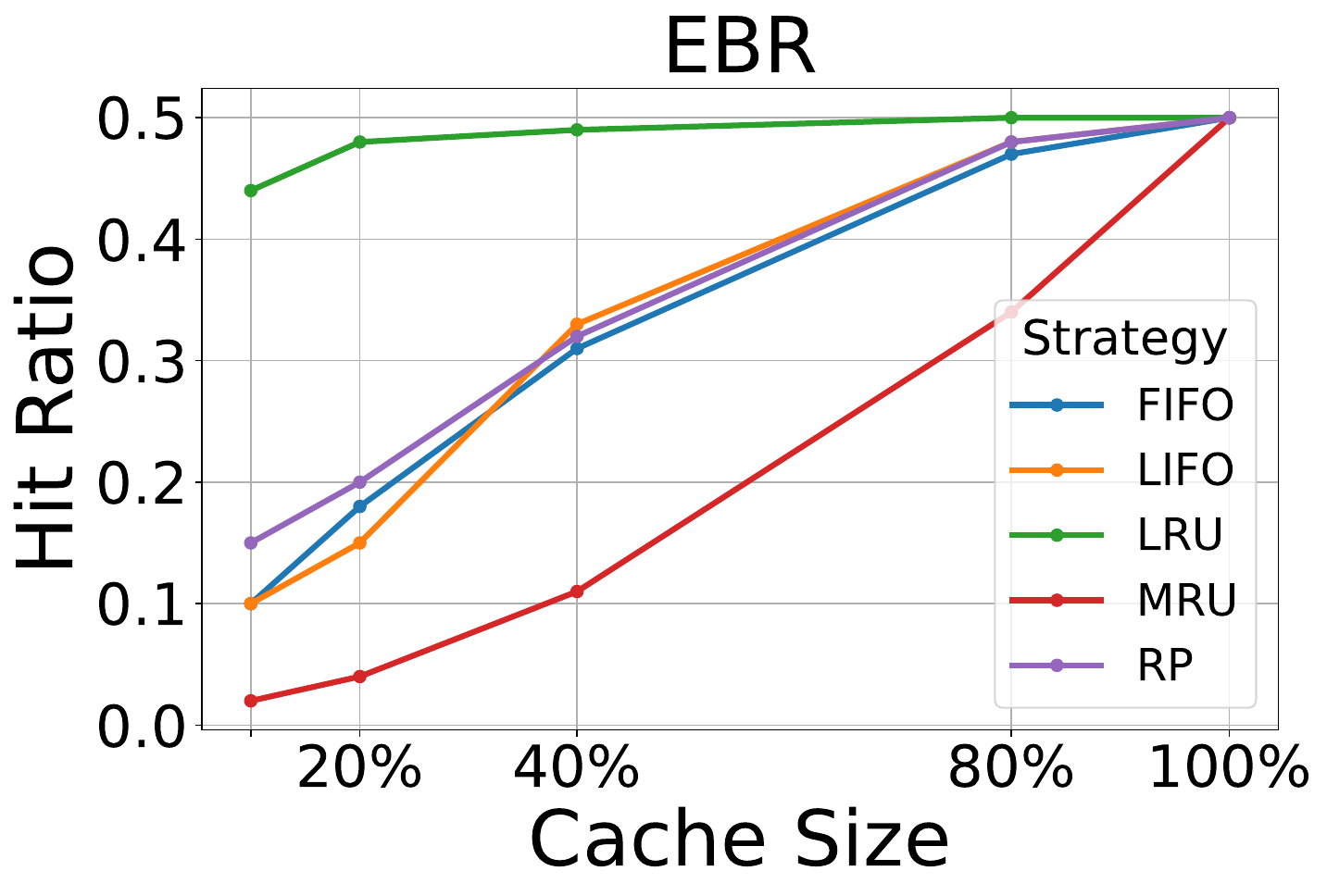}}
        \hfill
        \subcaptionbox*{}{\includegraphics[width=0.49\textwidth]{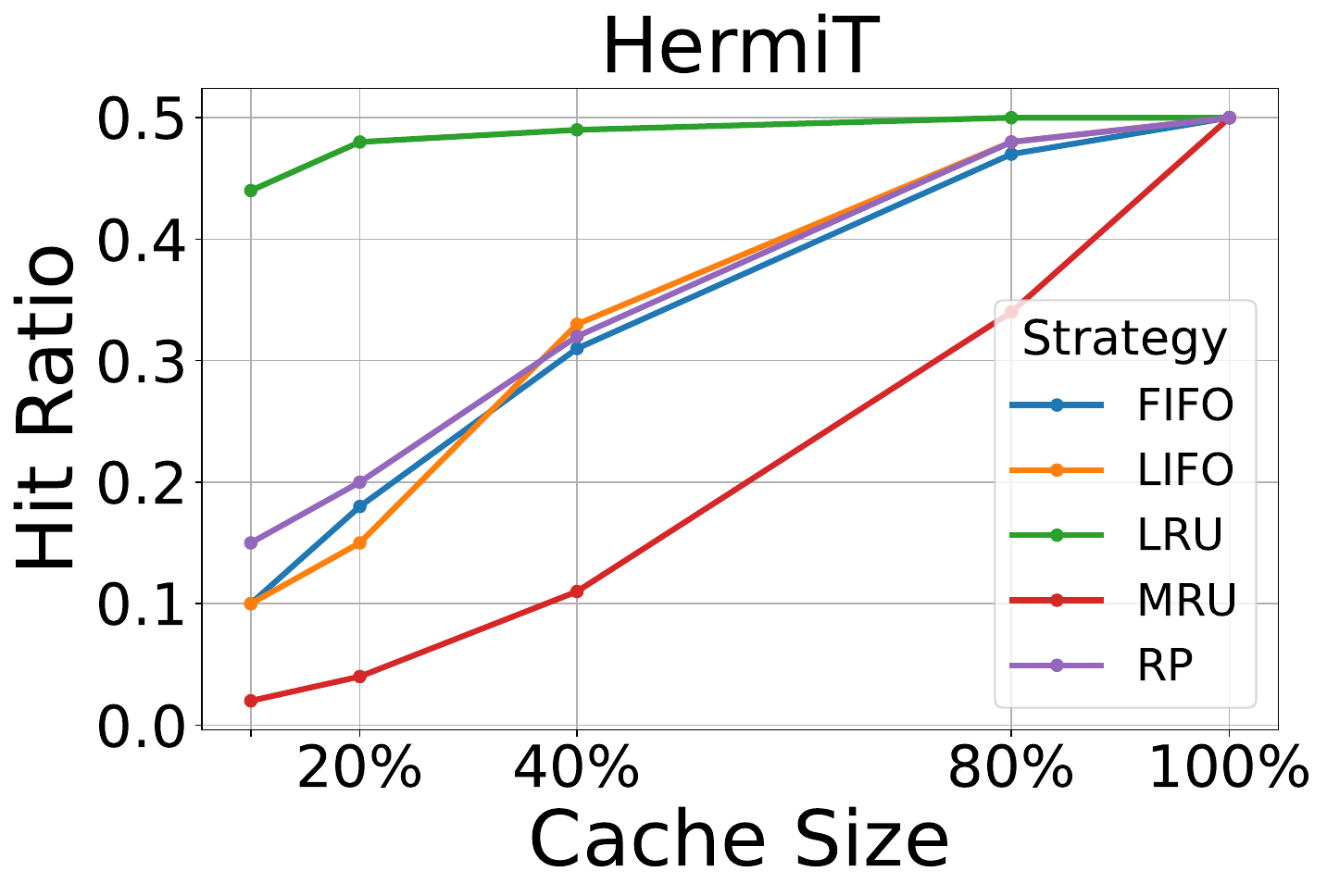}}
        
        \subcaptionbox*{}{\includegraphics[width=0.49\textwidth]{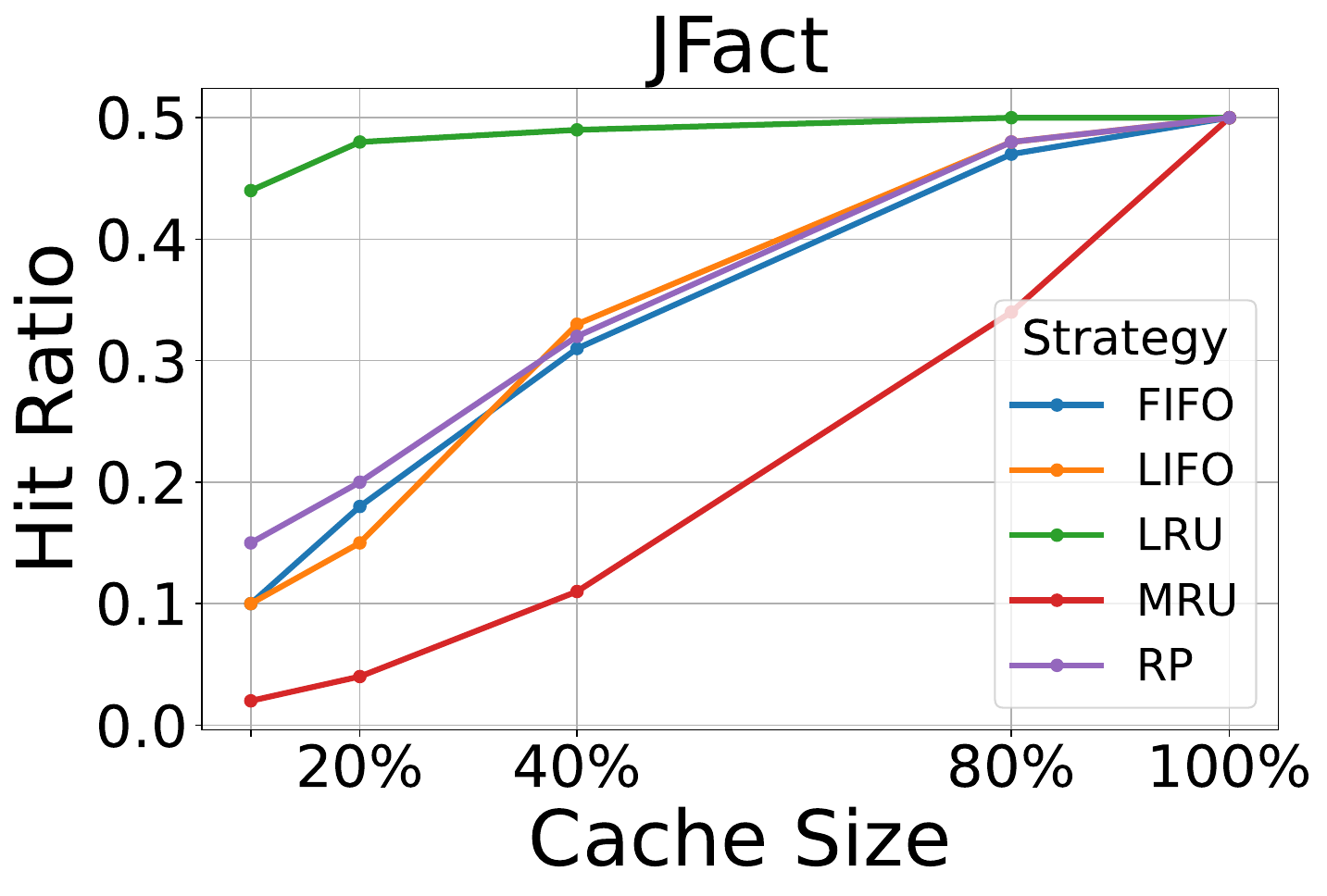}}
        \hfill
        \subcaptionbox*{}{\includegraphics[width=0.49\textwidth]{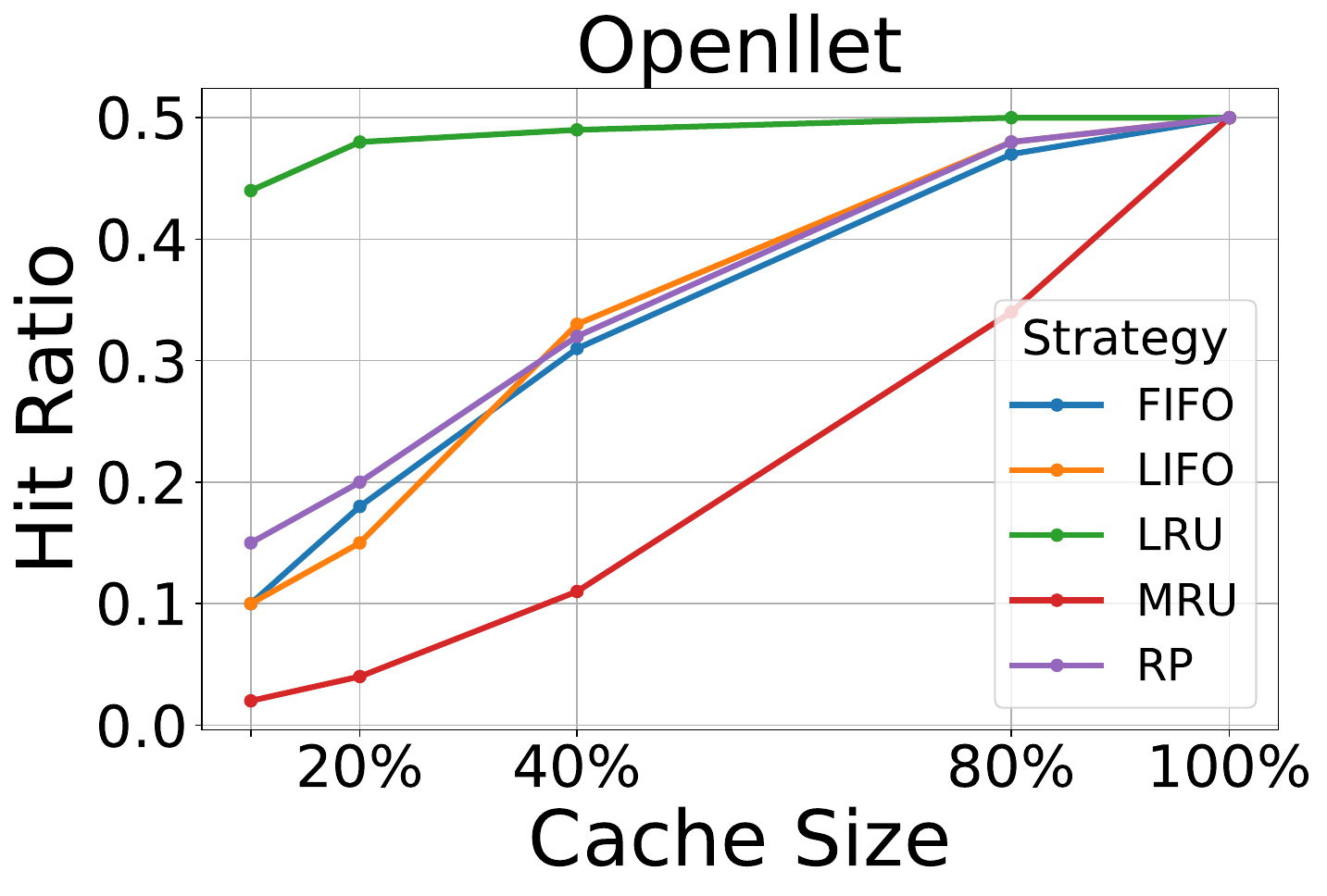}}
        \caption{Hit score performance vs cache size for each reasoner on the Family dataset without cache initialization. Higher values indicate more cache hits during retrieval.}
        \label{fig:result_hit_family_no_init}
    \end{minipage}
   
\end{figure*}

Figures \ref{fig:result_RT_family_init} and \ref{fig:result_RT_family_no_init} present the runtime results for concept retrieval on the Family dataset with and without cache initialization, respectively. The gray dotted line represents the baseline runtime of each reasoner without caching. Since the reasoner is executed independently for each cache size configuration, the baseline line is not perfectly flat and exhibits minor variations.

Across all reasoners, integrating the cache significantly reduces runtime, particularly as the cache size increases. Among the evaluated eviction strategies, LRU consistently provides the best performance. For instance, when the cache can store up to $80\%$ of the generated concepts, LRU yields the largest runtime reductions across all reasoners. In contrast, strategies such as RP and MRU show higher variability and less consistent improvements. LIFO also reduces runtime but generally remains slightly less effective than LRU.

Comparing warm and cold cache settings, the warm cache (initialized) leads to a more stable and faster decrease in runtime as the cache size grows. This improvement comes at the cost of a slightly higher runtime at the beginning of the execution due to the initialization phase. Overall, once the cache becomes populated, the warm configuration provides more consistent performance gains.

\paragraph{Hit ratio analysis}

Figures \ref{fig:result_hit_family_init} and \ref{fig:result_hit_family_no_init} show the hit ratio results on the Family dataset for the warm and cold cache settings, respectively. The other datasets exhibit similar trends.

As cache size increases, the hit ratio improves for all eviction strategies. LRU again demonstrates the most efficient behavior, achieving high hit ratios even at relatively small cache sizes (e.g., $20\%$ and $40\%$). In contrast, FIFO and MRU display slower improvements, indicating less efficient use of the available cache space. While the final hit ratios of warm and cold caches are similar at large cache sizes, the warm cache generally reaches high hit ratios faster and exhibits more stable behavior.

Overall, these results demonstrate that the proposed caching system substantially improves the performance of all evaluated reasoners. LRU consistently provides the best trade-off between runtime reduction and cache efficiency in both warm and cold cache settings. This observation aligns with the theoretical results of O’Neil et al.~\cite{o1999optimality}, which establish the optimality of LRU among common replacement policies. The results on the Family dataset, therefore, validate the effectiveness of our caching approach and motivate its evaluation on larger and more complex datasets.

\begin{figure}[htb]
    \centering
     \subcaptionbox{Carcinogenesis}{\includegraphics[width=0.44\linewidth]{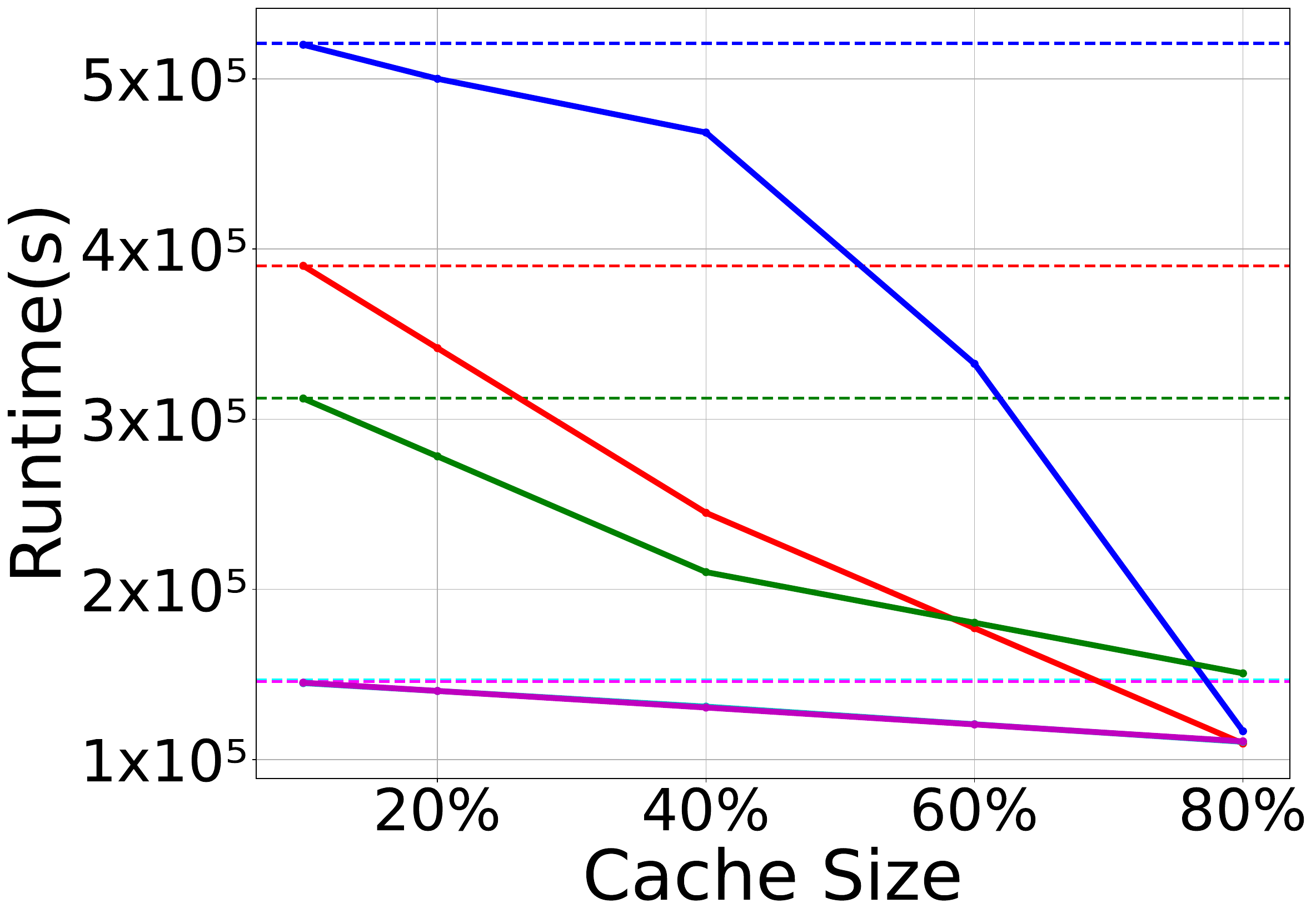}}
   \subcaptionbox{Mutagenesis}{\includegraphics[width=0.55\linewidth]{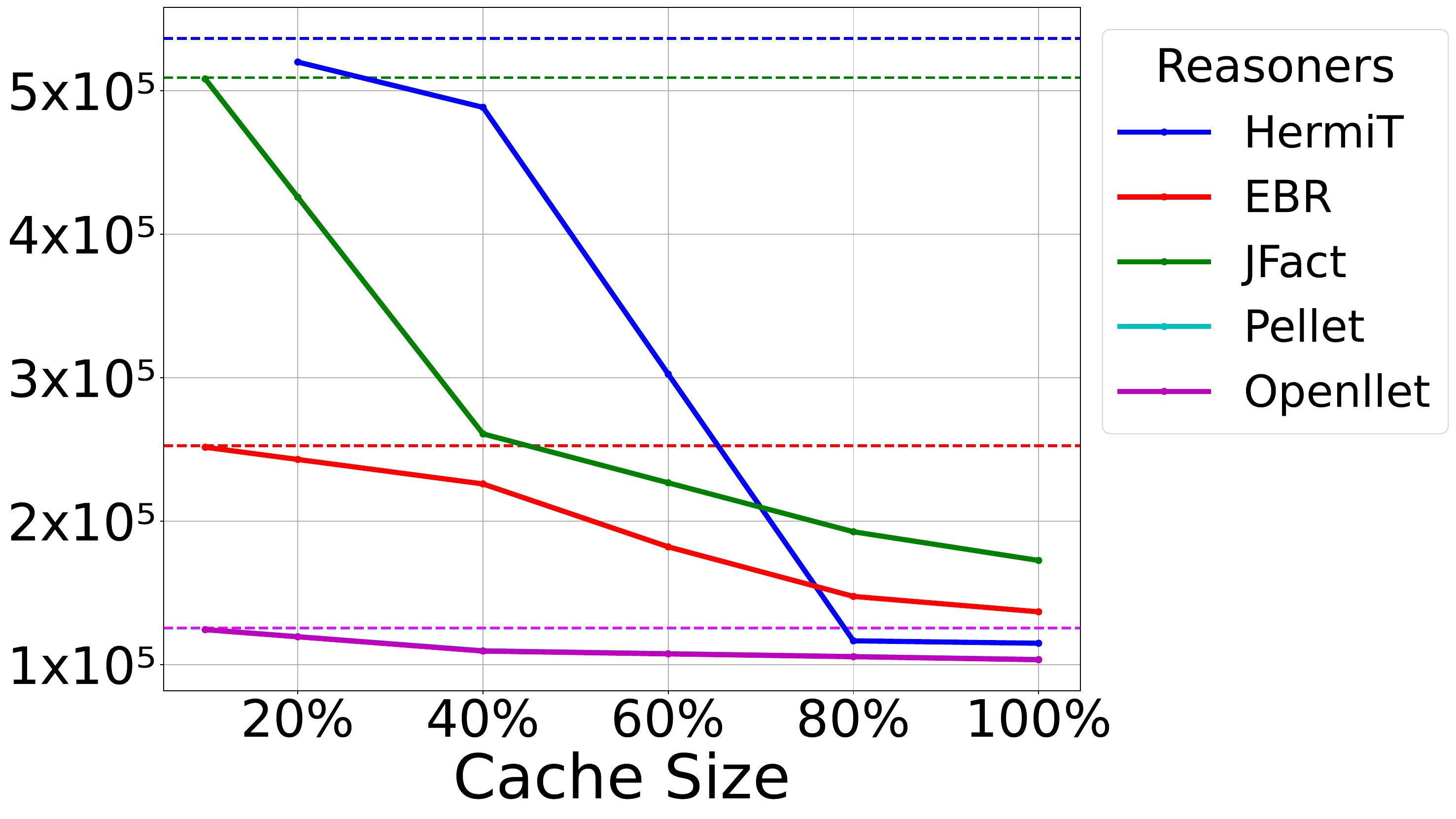}}
    \caption{Run time performance vs cache size for five reasoners on large datasets. The dotted horizontal lines represent the baseline performance of each reasoner without caching.}
    \label{fig:RT_large_data}
\end{figure}


Figure \ref{fig:RT_large_data} presents the runtime performance of all reasoners on the Carcinogenesis and Mutagenesis using the LRU eviction strategy. The results are shown as a function of cache size, ranging from $10\%$ to $100\%$. The dotted horizontal lines in each plot indicate the baseline runtime of the reasoners without caching. Across both datasets, the runtime consistently decreases as the cache size increases, highlighting the effectiveness of our caching mechanism in improving performance.

HermiT, the slowest reasoner without caching, benefits the most, with performance improvements reaching up to $60\%$ at larger cache sizes. Similarly, EBR, while faster than HermiT, shows runtime reductions of approximately $30\%$ to $50\%$, also at higher cache sizes. Pellet and Openllet show almost similar performance, with Openllet consistently maintaining its position as the fastest reasoner, both with and without caching. They achieve runtime reductions of approximately $45\% - 50\%$ at higher cache allocations. JFact sits between these reasoners in terms of baseline performance and acceleration, achieving runtime reductions of $25\% - 45\%$ across high cache sizes. Notably, the Carcinogenesis dataset shows slightly greater performance gains for most reasoners compared to the Mutagenesis dataset, reflecting the higher reasoning complexity of Carcinogenesis and the cache's ability to handle it effectively. For instance, the initial runtime of \approach was approximately 700,000 seconds, corresponding to more than eight days. However, with the integration of our cache, that runtime was drastically reduced to around 100,000 seconds, approximately equivalent to just a single day.
 The dotted baseline lines emphasize the advantage of integrating our cache, showcasing its ability to accelerate reasoning processes even for computationally demanding datasets.


\begin{figure}[ht]
    \centering  \subcaptionbox*{Carcinogenesis}
{\includegraphics[width=0.35\textwidth]{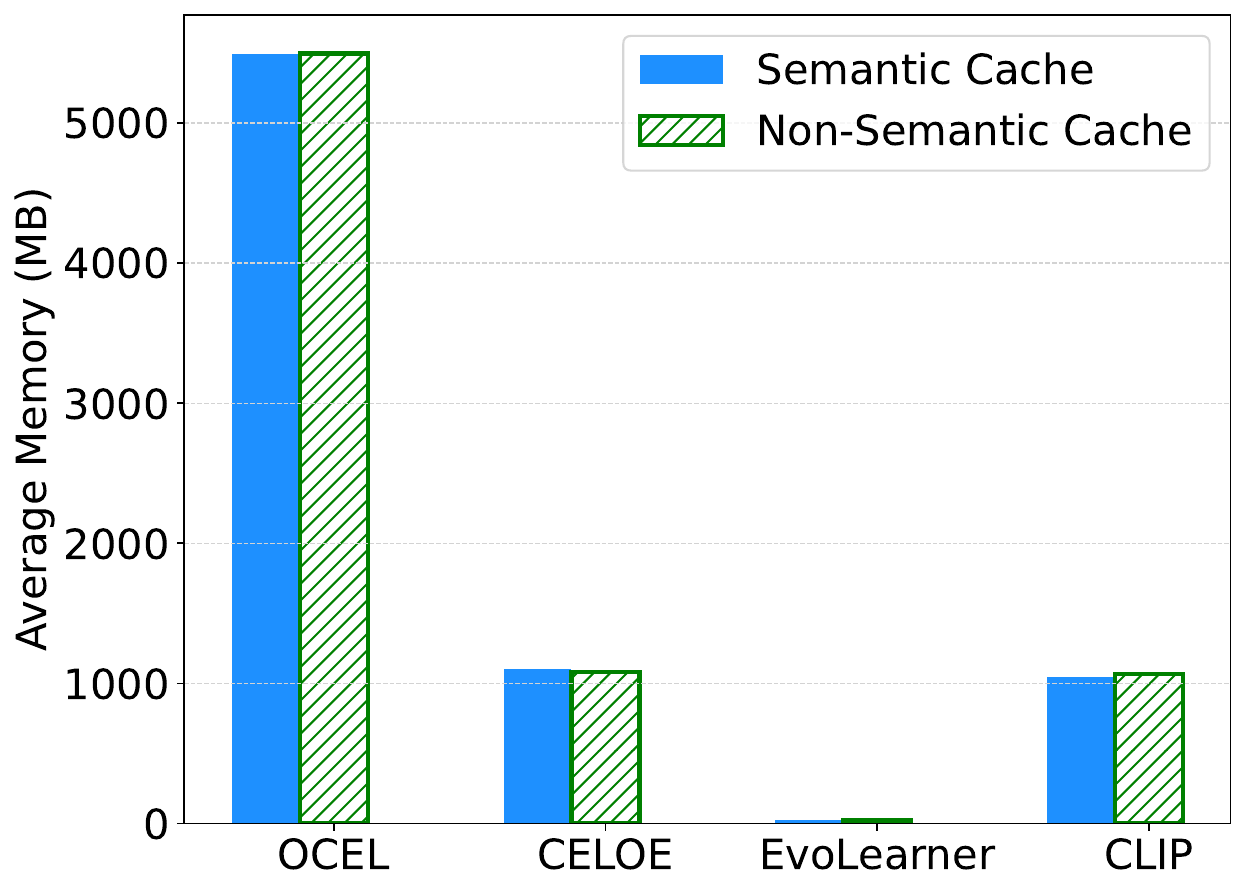}}
\subcaptionbox*{Vicodi}{\includegraphics[width=0.35\textwidth]{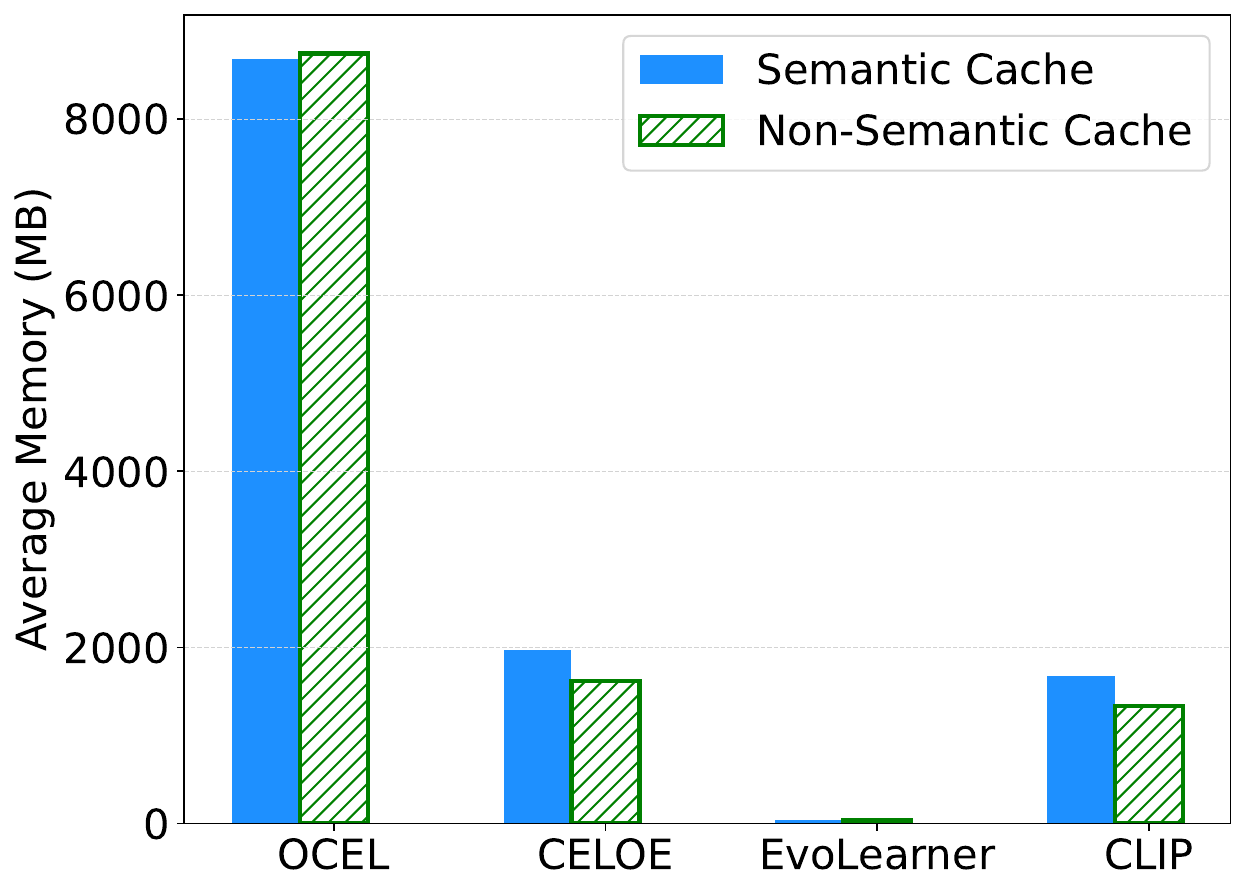}}
\subcaptionbox*{Mutagenesis}{\includegraphics[width=0.35\textwidth]{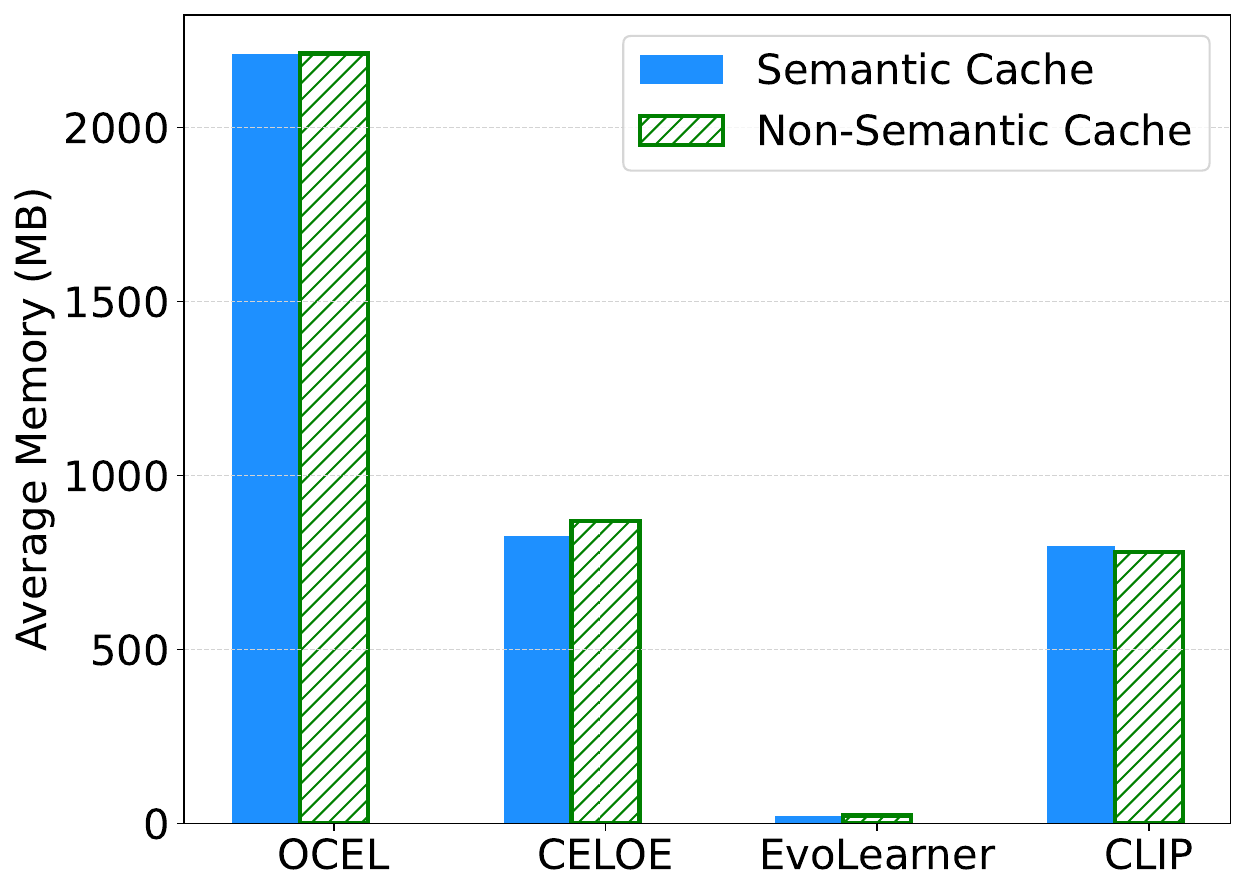}}
\subcaptionbox*{Family}{\includegraphics[width=0.35\textwidth]{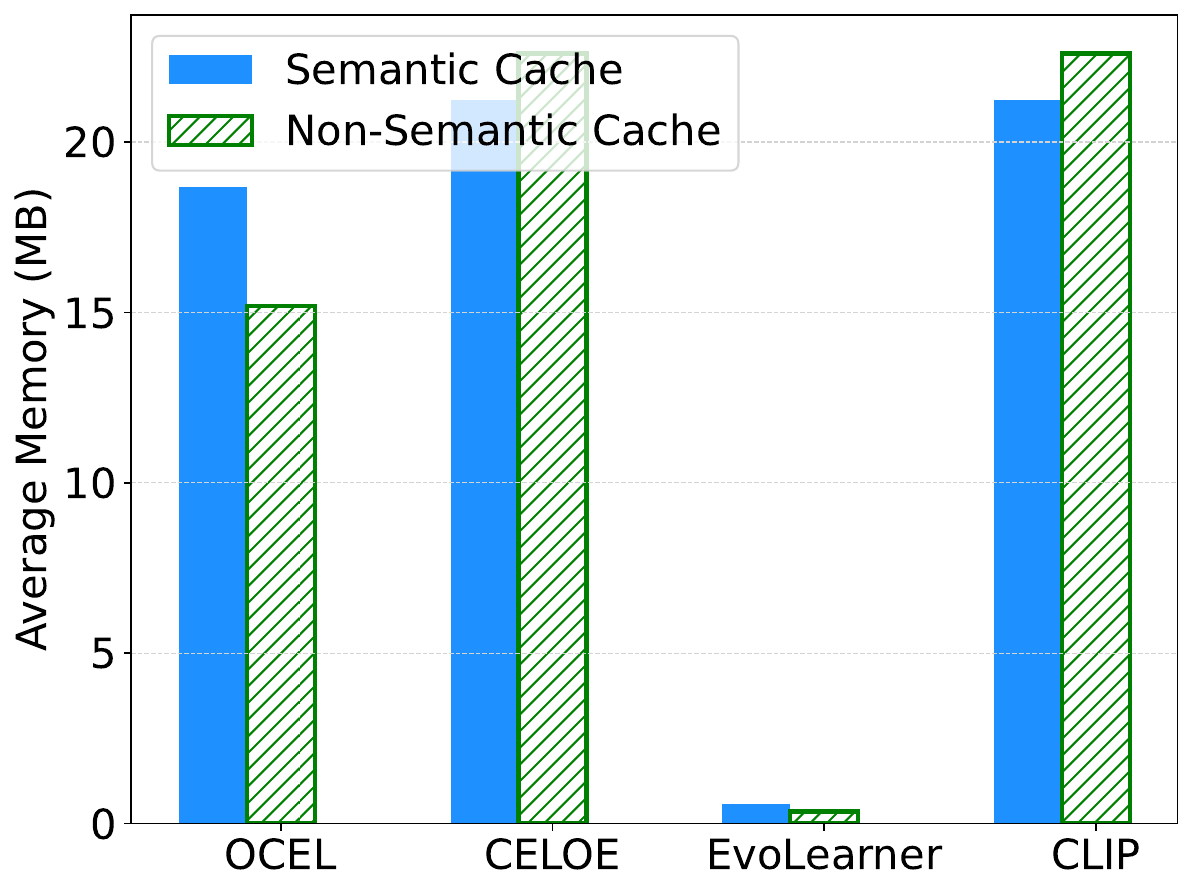}}
 \caption{Average physical memory usage of concept learners using the semantic cache, and the non-semantic cache baseline across different datasets.}
    \label{fig:cel_results_memory_cache}
\end{figure}

\subsection{Results on Class Expression Learning}

To evaluate the impact of our caching algorithm on CEL, we measured runtime performance across four datasets using several state-of-the-art learners. We considered three top-down refinement-based approaches—OCEL, CELOE, and CLIP—as well as EvoLearner, which follows an evolutionary strategy (see Section~\ref{sec:related_work}). Learning problems and corresponding positive and negative examples were generated following the protocol of \cite{demir2023neuro}. In all experiments, the maximum cache capacity was set to 1024 concepts.

The results in Figure~\ref{fig:cel_results_cache} show that our semantic cache substantially accelerates concept learning for refinement-based learners. The most pronounced improvements are observed for OCEL. For example, on the Carcinogenesis dataset, the average runtime decreases from more than 100 seconds without caching to under 20 seconds with the semantic cache. Similar reductions are observed on Mutagenesis and Vicodi. These gains are expected since refinement-based learners repeatedly perform concept retrieval operations, which our cache accelerates by reusing previously computed results.

CELOE and CLIP also benefit from the cache, although to a lesser extent. For instance, on Vicodi, the runtime of CELOE decreases from roughly 135 seconds to around 100 seconds. While these learners still rely on refinement operators, they typically generate fewer expensive retrieval queries, which explains the smaller but consistent improvements.

In contrast, the non-semantic cache provides little or no improvement and occasionally degrades performance. Because it stores concepts without considering their semantic structure, the cache quickly fills with syntactically distinct expressions that provide little opportunity for reuse. As a result, cache lookups rarely produce useful results and introduce additional overhead. This behavior highlights the importance of semantic awareness in enabling effective reuse.

EvoLearner does not benefit from the cache. Unlike refinement-based learners, it generates a large population of candidate concepts at the beginning of the search and evaluates them upfront. Consequently, concept retrieval is performed mostly once per concept, leaving little opportunity for reuse through caching.

Figure~\ref{fig:cel_results_memory_cache} reports the average physical memory consumption of the semantic and non-semantic caches across datasets. The results show that both variants exhibit similar memory usage, indicating that the semantic reasoning layer does not introduce significant additional overhead. Memory consumption mainly depends on the number of stored concepts and the size of their instance sets rather than on the semantic processing itself.

\section{Conclusion}

In this work, we proposed a simple yet effective caching algorithm to optimize the retrieval process, which is the most computationally intensive task in concept learning. By integrating our caching mechanism with various cache eviction strategies, we demonstrated its ability to enhance reasoners' performance significantly. Our results showed that the LRU eviction strategy works best with our cache, improving reasoning speed by at least $20\%$ and up to $60\%$ in some cases. This work highlights the importance of leveraging semantic caching to alleviate computational bottlenecks in ontology-based systems, providing a powerful yet straightforward solution for accelerating concept learning workflows.

\subsubsection*{Acknowledgements} 
This work was funded by the EU’s Horizon Europe programme (Marie Skłodowska-Curie grant No. 101073307) and supported by the Ministry of Culture and Science of North Rhine-Westphalia (MKW NRW) through the WHALE project (LFN 1-04) and the SAIL project (grant No. NW21-059D).

\clearpage

\bibliographystyle{splncs04}
\bibliography{lit}

\end{document}